%% file: main.tex
\documentclass[runningheads]{llncs}

\usepackage{eccv}

\usepackage{eccvabbrv}

\usepackage{graphicx}
\usepackage{booktabs}

\usepackage[accsupp]{axessibility}  
\usepackage{multirow}
\usepackage{wrapfig}
\usepackage{graphicx}

\usepackage{caption}
\usepackage{hyperref}

\usepackage{orcidlink}

\newcommand{\blankfootnote}[1]{%
  \begingroup
  \renewcommand\thefootnote{}\footnote{#1}%
  \addtocounter{footnote}{-1}%
  \endgroup
}

\begin{document}

\title{Textual Knowledge Matters: \\
Cross-Modality Co-Teaching for 
\\ Generalized Visual Class Discovery} 

\titlerunning{Cross-Modality Co-Teaching for
Generalized Visual Class Discovery}

\author{Haiyang Zheng\inst{1}\textsuperscript{$\star$}
\and
Nan Pu\inst{1}\textsuperscript{$\star$}
\and
Wenjing Li\inst{2,3}\textsuperscript{$\dagger$}
 \and
Nicu Sebe\inst{1}
\and
Zhun Zhong\inst{2,3}\textsuperscript{$\dagger$}
}

\authorrunning{H. Zheng et al.}

\institute{\textsuperscript{1}University of Trento \textsuperscript{2}Hefei University of Technology
\textsuperscript{3}University of Nottingham
\blankfootnote{\textsuperscript{$\star$} Equal contribution. \quad \textsuperscript{$\dagger$} Corresponding author.}
}

\maketitle

\begin{abstract}
In this paper, we study the problem of Generalized Category Discovery (GCD), which aims to cluster unlabeled data from both known and unknown categories using the knowledge of labeled data from known categories. Current GCD methods rely on only visual cues, which however neglect the multi-modality perceptive nature of human cognitive processes in discovering novel visual categories. To address this, we propose a two-phase TextGCD framework to accomplish multi-modality GCD by exploiting powerful Visual-Language Models. TextGCD mainly includes a retrieval-based text generation (RTG) phase and a cross-modality co-teaching (CCT) phase. First, RTG constructs a visual lexicon using category tags from diverse datasets and attributes from Large Language Models, generating descriptive texts for images in a retrieval manner. Second, CCT leverages disparities between textual and visual modalities to foster mutual learning, thereby enhancing visual GCD. In addition, we design an adaptive class aligning strategy to ensure the alignment of category perceptions between modalities as well as a soft-voting mechanism to integrate multi-modality cues. Experiments on eight datasets show the large superiority of our approach over state-of-the-art methods. Notably, our approach outperforms the best competitor, by 7.7\% and 10.8\% in All accuracy on ImageNet-1k and CUB, respectively. Code is available at \href{https://github.com/HaiyangZheng/TextGCD}{https://github.com/HaiyangZheng/TextGCD}.
\end{abstract}

\section{Introduction}
\label{sec:intro}
\begin{figure}[t]
  \centering
   \includegraphics[width=\linewidth]{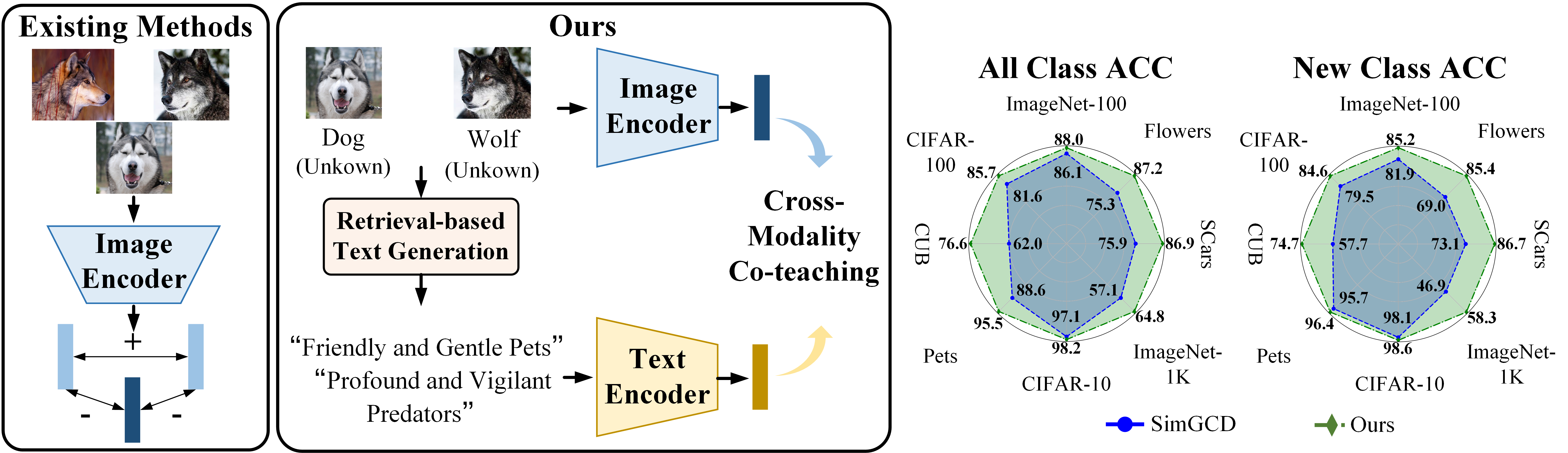}
   \caption{Left: Comparison to existing methods. Our approach introduces textual modality information (\textit{e.g.}, ``Friendly and Gentle Pets'' for Husky dogs, ``Profound and Vigilant Predators'' for wolves) into the framework and proposes cross-modal co-teaching for accurate generalized category discovery. Right: Performance comparison with SOTA.}
   \label{fig:highlight}
\end{figure}

Despite the remarkable advancements of deep learning in visual recognition, a notable criticism is that the models, once trained, show a significant limitation in recognizing novel classes not encountered during the supervised training phase. Drawing inspiration from the innate human capacity to seamlessly acquire new knowledge with reference to previously assimilated information, Generalized Category Discovery (GCD)~\cite{gcd} is proposed to leverage the knowledge of labeled data from known categories to automatically cluster unlabeled data that belong to both known and unknown categories. While current GCD methods~\cite{gcd,simgcd,dccl, gpc,promptcal} have demonstrated considerable success utilizing advanced large-scale visual models (\textit{e.g.}, ViT~\cite{vit}), as shown in Fig.~\ref{fig:highlight}, they predominantly focus on visual cues. In contrast, human cognitive processes for identifying novel visual categories usually incorporate multiple modalities~\cite{sloutsky2010perceptual}, such as encompassing visual, auditory, and textual elements in recognizing a subject. In the light of this, unlike existing GCD methods~\cite{gcd,simgcd,dccl, gpc,promptcal,sptnet} that rely on only visual modality, we propose to exploit both visual and textual cues for GCD. 

In this paper, we advocate the utilization of large-scale pre-trained Visual-Language Models (VLMs) (\textit{i.e.} CLIP~\cite{clip}) to inject rich textual information into GCD. However, VLMs require predetermined, informative textual descriptors (\textit{e.g.}, class names) for matching or recognizing images. This poses a significant challenge in GCD involving unlabeled data, particularly for unknown categories that lack predefined class names. Hence, how to provide relevant textual cues for unlabeled data is of a key in facilitating GCD with textual information.

\par

To this end, we introduce the TextGCD framework, comprising a retrieval-based text generation (RTG) phase and a cross-modality co-teaching (CCT) phase, to leverage the mutual benefit of visual and textual cues for solving GCD. In particular, RTG constructs a visual lexicon encompassing category tags from diverse datasets and attributes gleaned from Large Language Models (LLMs), which offer categorical insights. Subsequently, the most pertinent tags and attributes are selected from the lexicon to craft descriptive texts for all images through an offline procedure. While the augmentation of textual information for unlabeled images via RTG is beneficial, the effective integration of this textual information to enhance visual category discovery remains an essential challenge.

To solve the challenge, we introduce CCT to leverage the inherent disparities between textual and visual modalities, harnessing these distinctions to facilitate joint learning, which is important for co-teaching~\cite{coteaching}. To this end, we devise a cross-modal co-teaching training strategy that fosters mutual learning and collective enhancement between text and image models. Additionally, to ensure the effective alignment of category perceptions between text and image models and to facilitate comprehensive learning from both modalities, we introduce a warm-up stage and a class-aligning stage before engaging in cross-modality co-teaching. Finally, a soft-voting mechanism is deployed to amalgamate insights from both modalities, aiming to enhance the accuracy of category determination. In short, introducing the textual modality and exchanging insights between modalities have significantly improved the precision of visual category identification, particularly for previously unseen categories. We make the following contributions: 
\begin{itemize}
    \item We identify the limitations of existing GCD methods that rely on only visual cues and introduce additional textual information through a customized RTG based on large-scale VLMs.
    \item We propose a co-teaching strategy between textual and visual modalities, along with inter-modal information fusion, to fully exploit the strengths of different modalities in category discovery.
    \item Comprehensive experiments on eight datasets demonstrate the effectiveness of the proposed method. Notably, compared to the leading competitor in terms of All accuracy, our approach achieves an increase of 7.7\%  on ImageNet-1k and 10.8\% on CUB. The source code will be publicly available.
\end{itemize}

\section{Related Works}
\noindent \textbf{Generalized Category Discovery (GCD)} aims to accurately classify an unlabeled set containing both known and unknown categories, based on another dataset labeled with only known categories. It is an extension of novel category discovery~\cite{zhong2021openmix,roy2022class,han2021autonovel,han2019learning,fini2021unified,zhong2021neighborhood} where the unlabeled set only contains unknown classes. Vaze et al.~\cite{gcd} first proposed optimizing image feature similarities through supervised and self-supervised contrastive learning to address GCD. Building on this, SimGCD~\cite{simgcd} designed a parametric classification baseline for GCD. For better image representations, DCCL~\cite{dccl} proposed a novel approach to contrastive learning at both the concept and instance levels. Similarly focusing on image representation, PromptCAL~\cite{promptcal} introduced a Contrastive Affinity Learning approach with auxiliary visual prompts designed to amplify the semantic discriminative power of the pre-trained backbone. Furthermore, SPTNet~\cite{sptnet} proposed iteratively implementing model finetuning and prompt learning, resulting in clearer boundaries between different semantic categories. 
Unlike existing methods that primarily focus on visual cues, we design a retrieval-based approach to introduce text cues from LLM for unknown categories. Recently, CLIP-GCD~\cite{clipgcd} concatenated image features with textual features obtained from a Knowledge Database and categorized them using a clustering method. Unlike CLIP-GCD, which merely concatenates text and visual features from a frozen backbone, we leverage the differences between textual and visual models to establish a dynamic co-teaching scheme.

\par\noindent
\textbf{Visual-Language Models (VLMs)} are designed to map images and text into a unified embedding space, facilitating cross-modal alignment. Among the prominent works, CLIP~\cite{clip} employs contrastive representation learning with extensive image-text pairs, showcasing remarkable zero-shot transfer capabilities across a variety of downstream tasks. Additionally, LENS~\cite{lens} leverages VLMs as visual reasoning modules, integrating them with Large Language Models (LLMs) for diverse visual applications. As VLMs effectively bridge the visual and textual modalities, enhancing image classification tasks with knowledge from LLMs has been extensively explored~\cite{vlm_for_cls1, vlm_for_cls2, vlm_for_cls3}. However, these works use VLMs to enhance image classification in scenarios where the names of all classes are predefined for VLMs. In this paper, we exploit VLMs to address the GCD, where the unlabeled data do not have exploitable textual information for VLMs. We utilize a retrieval-based text generation method to enable the GCD to benefit from VLMs.

\noindent 
\textbf{Co-teaching}, which stemmed from the Co-training approach~\cite{co-training}, was originally proposed as a learning paradigm to address the issue of label noise,~\cite{coteaching}. This strategy involves training two peer networks that select low-loss instances from a noisy mini-batch to train each other. Yuan et al.~\cite{coteaching_for_OMO} established three symmetric peer proxies with pseudo-label-driven co-teaching to address the Offline model-based optimization task. Yang et al.~\cite{coteaching_for_reid1} were the first to introduce the co-teaching strategy into the Person re-identification task, designing an asymmetric co-teaching framework. Similarly, Roy et al.~\cite{coteaching_for_MDA} pioneered co-teaching between classifiers of a dual classifier head to tackle the multi-target domain task. Unlike these works, our focus is on applying co-teaching to the GCD task. Furthermore, rather than employing two peer networks within the same modality or learning from self-discrepancy, we initiate co-teaching between the image model and the text model, leveraging the disparities between modalities to foster mutual enhancement.

\section{Method}
\label{sec:method}
\noindent \textbf{Task Configuration:} 
In GCD, we are given a labeled dataset, denoted as $\mathcal{D}_L=\{(\mathbf{x}_i,y_i^l)\}_{i=1}^M\subseteq\mathcal{X}\times\mathcal{Y}_L$, and an unlabeled dataset, denoted as $\mathcal{D}_U=\{(\mathbf{x}_i,y_i^u)\}_{i=1}^N\subseteq\mathcal{X}\times\mathcal{Y}_U$, where $M$ and $N$ indicate the number of samples in the $\mathcal{D}_L$ and $\mathcal{D}_U$, respectively. $\mathcal{Y}_L$ and $\mathcal{Y}_U$ indicate the label spaces for labeled and unlabeled datasets, respectively, where $\mathcal{Y}_L\subseteq\mathcal{Y}_U$. That is, $\mathcal{D}_U$ contains data from unknown categories, and $\mathcal{Y}_U$ is unavailable. The objective of GCD is to leverage the prior knowledge from known categories within $\mathcal{D}_L$ to classify samples in $\mathcal{D}_U$ effectively. Following~\cite{simgcd}, it is assumed that the number of classes in $\mathcal{D}_U$, represented by $K=|{\mathcal Y}_U|$, is predetermined.

\subsection{Overview}
\begin{figure*}[ht]  
  \centering  \includegraphics[width=0.9\linewidth]{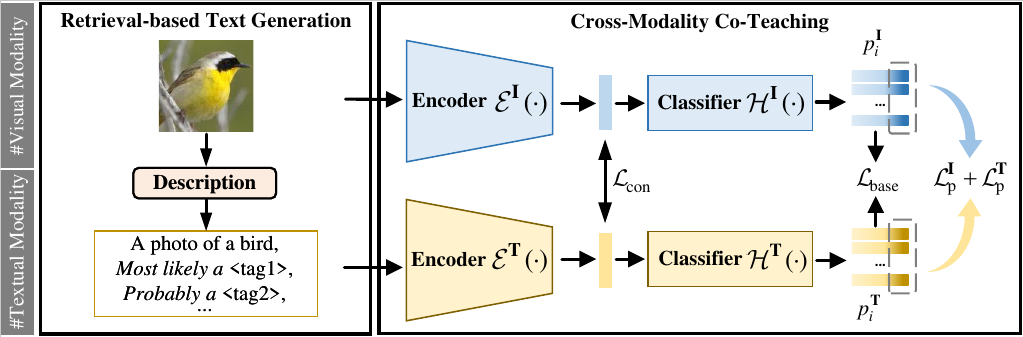}  
  \caption{The TextGCD framework comprises two main phases: Retrieval-based Text Generation (RTG) and Cross-modality Co-Teaching (CCT). In the RTG phase, descriptions for each sample are generated using a visual lexicon. The CCT phase involves developing a two-stream parametric model that leverages the interaction of visual and textual modalities for enhanced mutual progress. The gray dashed box on the right illustrates the text and image models independently selecting high-confidence samples with pseudo labels for the co-teaching process.}
  \label{fig:framework}
\end{figure*}

The framework of the proposed TextGCD, depicted in Fig.~\ref{fig:framework}, consists of the Retrieval-based Text Generation (RTG) phase and Cross-modal Co-Teaching (CCT) phase. In RTG, we initially develop a visual lexicon comprising a broad spectrum of tags, along with attributes obtained from LLMs. Subsequently, we extract representations of images and the visual lexicon with an auxiliary VLM model and generate textual category information for each image in a retrieval manner. Building upon this, we design textual and visual parametric classifiers and propose the CCT phase to foster mutual learning and collective progress between text and image models.

\subsection{Baseline}
\label{sec:baseline}
We follow the SimGCD~\cite{simgcd} to build our parametric learning baseline, which consists of a supervised loss for labeled data and an unsupervised loss for both labeled and unlabeled data. 

Specifically, the supervised loss utilizes the conventional cross-entropy loss:
\begin{equation}
\mathcal{L}_{\text{sup}}=\frac{1}{|B^{l}|}\sum_{i\in B^{l}}\ell(\boldsymbol{y}_{i},\boldsymbol{p}_{i}),
  \label{eq:loss_sup_I}
\end{equation}
where $B^{l}$ indicates the mini-batch for labeled data, $\ell$ is the cross-entropy loss, and $\boldsymbol{p}_{i}=\sigma({\mathcal{H}(\mathcal{E}(\mathbf{x}_i))/\tau})$ is the predicted probabilities of input $\mathbf{x}_i$. Here, $\mathcal{E}$ represents the backbone encoder, and $\mathcal{H}$ indicates the parametric classifier. $\sigma(\cdot)$ denotes the softmax function, and $\tau$ is a temperature parameter set to $\tau_s$.

For unsupervised loss, we use the predicted probabilities of an augmented counterpart $\mathbf{x}_i^{\prime}$ as the supervision to calculate the classification loss for the original input $\mathbf{x}_i$, which is formulated as:
\begin{equation}
\mathcal{L}_{\text{unsup}}=\frac{1}{|B|}\sum_{i\in B}\ell({\boldsymbol{q}^{\prime}_{i}},\boldsymbol{p}_{i})-\varepsilon H(\overline{\boldsymbol{p}}),
  \label{eq:loss_unsup_I}
\end{equation}
where $\boldsymbol{q}^{\prime}_{i}$ is the predicted probabilities of $\mathbf{x}_i^{\prime}$ using a sharper temperature value $\tau_t$. We also include a mean-entropy maximization regularizer $H(\overline{\boldsymbol{p}})$~\cite{assran2022masked} for the unsupervised objective. $H(\overline{\boldsymbol{p}})$=$-\sum_{k}\overline{\boldsymbol{p}}^{(k)}\log\overline{\boldsymbol{p}}^{(k)}$, where $\overline{\boldsymbol{p}}=\frac{1}{2|B|}\sum_{i\in B}(\boldsymbol{p}_{i}+\boldsymbol{p}_{i}^{\prime})$. $\boldsymbol{p}_{i}$ and $\boldsymbol{p}_{i}^{\prime}$ are the probabilities of $\mathbf{x}_i$ and $\mathbf{x}_i^{\prime}$, respectively, which use the same temperature of $\tau_u$. $B$ indicates the mini-batch for both labeled and unlabeled data. The hyperparameter $\varepsilon$ aligns with the configuration used in SimGCD~\cite{simgcd}.

The classifier is jointly trained with supervised loss and unsupervised loss, formulated as:
\begin{equation}
\mathcal{L}_{\text{base}}=\lambda \cdot \mathcal{L}_{\text{sup}}+(1-\lambda)\cdot\mathcal{L}_{\text{unsup}},
  \label{eq:loss_cls}
\end{equation}
where $\lambda$ serves as the balancing factor.

\subsection{Retrieval-based Text Generation}
\label{sec:text generation}
\begin{wrapfigure}[11]{r}{0.58\textwidth}
  \centering
   \includegraphics[width=0.55\textwidth]{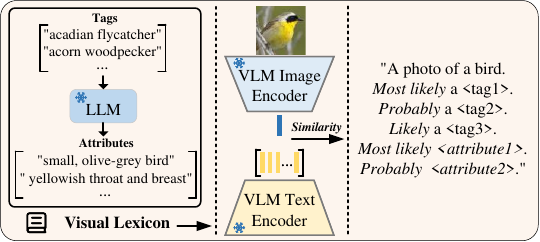}
   \caption{Schema of retrieval-based text generation. }
   \label{fig:lens}
\end{wrapfigure}

The existing foundational VLMs and LLMs, trained on massive amounts of data, have demonstrated remarkable capabilities in aligning visual images with corresponding textual descriptions and generating detailed object descriptions. Drawing inspiration from these advancements, we introduce the Retrieval-based Text Generation (RTG) phase for GCD. This process involves two stages: constructing a comprehensive visual lexicon and retrieving probable tags from this lexicon to generate textual information for images.

\noindent\textbf{Building the Visual Lexicon}. To construct a visual lexicon, we aggregate a diverse and extensive collection of tags from established benchmarks in semantic segmentation, object detection and classification. These benchmarks are selected for their comprehensive coverage of the visual categories in the world. Additionally, we utilize a large language model (\textit{e.g.}, GPT3~\cite{gpt3}) to further describe each tag by using the prompt of ``What are useful features for distinguishing a \{tag\} in a photo?'', which can augment the visual lexicon with the attribute of ``\{tag\} which has the \{feature\}''.

\noindent\textbf{Tag Retrieval}. Given the visual lexicon, comprising tags and corresponding attributes, we utilize an auxiliary VLM (\textit{e.g.}, CLIP~\cite{clip}) to encode it, thereby creating a textual feature bank. For each image, we use the auxiliary model to generate its visual feature, which is then compared against the entries in the textual feature bank using cosine similarity. Subsequently, we identify the most relevant $n_t$ tags and $n_a$ attributes exhibiting the highest similarity to the image. We then concatenate the textual descriptors of these tags and attributes based on the similarity ranking. This process yields the categorical descriptive text $\mathbf{t}_i$ for each image $\mathbf{x}_i$. The workflow of the RTG phase is delineated in Fig.~\ref{fig:lens}.

\noindent\textbf{Discussion}. In this paper, we aim to harness the capabilities of the existing foundational models to deliver comprehensive and accurate textual information for GCD. To this end, we employ GPT-3~\cite{gpt3} as the LLM to obtain attributes and the ViT-H-based CLIP~\cite{clip} as the auxiliary VLM to identify tags and attributes. We also attempt to use a smaller auxiliary VLM, \textit{e.g.,} ViT-B-based CLIP, to generate the descriptive text. 
However, it demonstrated limited efficacy for fine-grained classification because ViT-B-based CLIP tends to focus on more general Tags and Attributes in the Visual Lexicon. Using Flower102 for instance, ViT-B-based CLIP tends to give high similarity to broader Tags like ``Plant \& Flower''.
Compared to existing methods that only use visual cues for GCD, we explore the possibility of introducing the textual information to facilitate the GCD task by using the freely available, public foundational models, which would be a new trend in the community. In addition, our approach enjoys several advantages. Firstly, the RTG is an offline phase, where we only need to process each sample once, avoiding the need for excessive computational overhead. Secondly, given the generated descriptive text, we only need to train the GCD model with a smaller model (ViT-B-based CLIP), achieving significantly higher results than existing methods. Third, the auxiliary VLM can be replaced with other advanced foundational models, such as FLIP~\cite{flip} and CoCa~\cite{coca}. Consequently, the effectiveness of our method is likely to be enhanced in line with the advancements in these foundational models.

\subsection{Cross-modality Co-teaching}
\label{sec:cross-modality co-teaching}

In light of the categorical descriptive texts generated during the RTG phase, the primary challenge lies in effectively utilizing them to train an accurate GCD model. 
To address this, we develop two-stream parametric classifiers for visual and textual modalities. Recognizing that the intrinsic differences between the modalities naturally meet the requirements for the model disparity in co-teaching strategies~\cite{coteaching}, we propose a Cross-modal Co-Teaching (CCT) phase to realize the mutual benefits of the visual and textual classifiers. However, a notable challenge arises due to the lack of annotations during the co-teaching process: the classifiers for each modality are often misaligned. This misalignment implies that consistent class indexes across the two modalities cannot be guaranteed, posing a significant obstacle to effective co-teaching. We introduce two preliminary stages to overcome this issue: a warm-up stage and a class-aligning stage, designed to establish modality-specific classifiers while ensuring their alignment. 

\noindent\textbf{Basic Loss}.
Given the inputs from two modalities, $\mathbf{x}_i$ and $\mathbf{t}_i$, we construct a text model and an image model based on the parametric baseline in Sec.~\ref{sec:baseline}. Thus, the basic loss for our method is formulated as:
\begin{equation}
\mathcal{L}_{\text{base}}=\mathcal{L}^{\mathbf I}_{\text{base}} + \mathcal{L}^{\mathbf T}_{\text{base}},
  \label{eq:loss_basic}
\end{equation}
where both components are implemented by Eq.~\ref{eq:loss_cls}. The difference is that $\mathcal{L}^{\mathbf T}_{\text{base}}$ is calculated by the text model and textual description.

\begin{wrapfigure}[12]{r}{0.55\textwidth}
  \centering
   \includegraphics[width=0.55\textwidth]{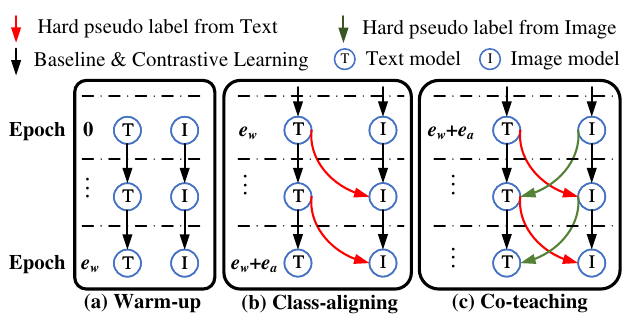}

   \caption{Schema of cross-modality co-teaching. }
   \label{fig:coteaching}
\end{wrapfigure}

\noindent\textbf{Image-Text Contrastive Learning}. To strengthen the association between the image and text modalities, we further introduce a cross-modal contrastive learning between image features and textual features, which is formulated as:
\begin{equation}
\mathcal{L}_{\text{con}}=\frac{1}{|B|}\sum_{i\in B}-\log\frac{\exp\left(\mathbf{f}_i^{\mathbf I}\cdot(\mathbf{f}_i^{\mathbf T})^\top/\tau_c\right)}{\sum_j\mathbb{I}_{[j\neq i]}\exp\left(\mathbf{f}_i^{\mathbf I}\cdot(\mathbf{f}_j^{\mathbf T})^\top/\tau_c\right)},
  \label{eq:conloss}
\end{equation}
where $\tau_c$ is a temperature parameter. $\mathbf{f}_i^{\mathbf I}=\mathcal{E}^{\mathbf I}(\mathbf{x}_i)$ is the image feature and $\mathbf{f}_i^{\mathbf T}=\mathcal{E}^{\mathbf T}(\mathbf{t}_i)$ is the textual feature. $\mathcal{E}^{\mathbf I}$ and $\mathcal{E}^{\mathbf T}$ represent the image and text encoders, respectively. We next introduce the three stages of the proposed CCT (see Fig.~\ref{fig:coteaching}), \textit{i.e.,} warm-up, class-aligning, and co-teaching.

\noindent\textbf{Stage I: Warm-up}. The image and text models initially undergo a warm-up training stage with $e_w$ epochs, as depicted in Fig~\ref{fig:coteaching}(a). During this stage, both the text model and image model are trained using the basic loss (Eq.~\ref{eq:loss_basic}) and cross-modal contrastive loss (Eq.~\ref{eq:conloss}). This warm-up training aims to enable both models to adequately learn category knowledge from their respective modality data, thereby developing modality-specific category perceptions.

\noindent\textbf{Stage II: Class-aligning}. Subsequently, to rectify the class misalignment between image and text classifiers, a class-aligning stage with $e_a$ epochs is introduced, as illustrated in Fig~\ref{fig:coteaching}(b). This stage aims to align the classifiers of two modalities. To this end, we select high-confidence samples based on the text model and use them to guide the training of the image model. Specifically, for each category $k$, we select the top $s$ samples that have the highest probabilities in the $k$-th category produced by the text model, formulated as $\operatorname{Top}_k^{\mathbf T}(s)$. In addition, for each sample $i$ in $\operatorname{Top}_k^{\mathbf T}(s)$, we assign it with the hard pseudo label of $\hat{y}_i^{\mathbf T}=k$. Thus, besides the basic and contrastive losses, we additionally train the image model by the pseudo-labeling loss:
\begin{equation}
\mathcal{L}_{\text{p}}^{\mathbf I}=\frac{1}{|B^{s}|}\sum_{i\in B^{s}}\ell(\hat{y}_i^{\mathbf T},\boldsymbol{p}_{i}^{\mathbf I}),
  \label{eq:loss_pseudo_I}
\end{equation}
where $|B^{s}|$ indicates the number of selected samples in the mini-batch.

\noindent\textbf{Stage III: Co-teaching}. Following the warm-up and class-aligning stages, we establish cross-modality co-teaching to facilitate the mutual benefit between the image and text models, as shown in Fig~\ref{fig:coteaching}(c). Similar to the class-aligning stage, we also generate high-confidence samples from the image model, which is defined as $\operatorname{Top}_k^{\mathbf I}(s)$. For each sample $i$ in $\operatorname{Top}_k^{\mathbf I}(s)$, the hard pseudo label is $\hat{y}_i^{\mathbf I}=k$, and thus the pseudo-labeling loss for the text classifier is formulated as:
\begin{equation}
\mathcal{L}_{\text{p}}^{\mathbf T}=\frac{1}{|B^{s}|}\sum_{i\in B^{s}}\ell(\hat{y}_i^{\mathbf I},\boldsymbol{p}_{i}^{\mathbf T}).
  \label{eq:loss_pseudo_T}
\end{equation}
The total loss of Stage III is formulated as:
\begin{equation}
\mathcal{L}=\mathcal{L}_{\text{base}}+\mathcal{L}_{\text{con}}+\mathcal{L}_{\text{p}}^{\mathbf I}+\mathcal{L}_{\text{p}}^{\mathbf T}.
  \label{eq:total_loss}
\end{equation}

\noindent\textbf{Inference}. When determining the category of object $i$, we employ soft voting to merge the category perceptions from both modalities:
 \begin{equation}
\boldsymbol{P}_{i}=  \boldsymbol{p}_{i}^{\mathbf I} + \boldsymbol{p}_{i}^{\mathbf T}.
  \label{eq:P_i}
\end{equation}
The predicted category is the one that achieves the largest value in $\boldsymbol{P}_{i}$, \textit{i.e.,} $\arg\max \boldsymbol{P}_{i}$. This integrated approach leverages the strengths of both modalities, leading to more accurate classification.

\section{Experiments}
\noindent \textbf{Datasets.} We conduct comprehensive experiments on four generic image classification datasets, \textit{i.e.}, CIFAR-10, CIFAR-100~\cite{cifar}, ImageNet-100, and ImageNet-1K~\cite{imagenet}, as well as on four fine-grained image classification datasets, \textit{i.e.}, CUB~\cite{cub}, Stanford Cars~\cite{scars}, Oxford Pets~\cite{pets}, and Flowers102~\cite{flowers}. Following~\cite{gcd}, for each dataset, we select half of the classes as the known classes and the remaining as the unseen classes. We select 50\% of samples from the known classes as the labeled dataset $\mathcal{D}_L$ and regard the remaining samples as unlabeled dataset $\mathcal{D}_U$ that are from both known and unseen classes.


\noindent\textbf{Evaluation Protocol.} Following \cite{gcd}, we use the clustering accuracy (ACC) to evaluate the performance of each algorithm. Specifically, the ACC is calculated on the unlabeled data, formulated as $ACC = \frac{1}{|\mathcal{D}_U|} \sum_{i=1}^{|\mathcal{D}_U|} \mathbb{I} (y_i = C(\overline{y}_i))$, where $\overline{y} =\arg\max \boldsymbol P$ represents the predicted labels and $y$ denotes the ground truth. The $C$ denotes the optimal permutation.

\noindent \textbf{Implementation Details.} We construct the tag lexicon by incorporating tags from semantic segmentation and object detection datasets~\cite{lvis,coco,openimages}, image classification datasets~\cite{imagenetlarge,Caltech101,wild,pets,food,sundatabase,3dobject} and visual genome dataset~\cite{visualgenome}. The CLIP~\cite{clip} model with a ViT-B-16 architecture serves as the backbone, ensuring a fair comparison to methods utilizing the DINO~\cite{dino} model with the same architecture. A linear classifier layer is added after the backbone for each modality. During training, only the last layers of both the text and image encoders in the backbone are fine-tuned, using a learning rate of 0.0005. The learning rates for classifiers start at 0.1 and decrease following a cosine annealing schedule. We train the model for 200 epochs on all datasets, using a batch size of 128 and processing dual views of randomly augmented images and texts. The temperature parameters $\tau_s$ and $\tau_u$ are set at 0.1 and 0.05, respectively. $\tau_t$ starts at 0.035 and decreases to 0.02, except for CIFAR-10 where it varies from 0.07 to 0.04 due to fewer classes. We use the logit scale value to set $\tau_c$ as in the CLIP model. The balancing factor $\lambda$ is set to 0.2. Warm-up and class-aligning stages last for 10 and 5 epochs, respectively, denoted as $e_w=10$ and $e_a=5$. For selecting high-confidence samples, we choose $s=r*\frac{|\mathcal{D}_U|}{K}$ samples per category, with $r$ set at 0.6. The average result over the three runs is reported.


\subsection{Comparison with the State of the Art}

We conduct a comparative analysis between our proposed TextGCD and leading methods in GCD, as delineated in Tab.~\ref{tab:comparison_generic} and Tab.~\ref{tab:comparison_finegtained}. These methods encompass GCD~\cite{gcd}, SimGCD~\cite{simgcd}, DCCL~\cite{dccl}, GPC~\cite{gpc}, PromptCAL~\cite{promptcal}, CLIP-GCD~\cite{clipgcd} and SPTNet~\cite{sptnet}. Except for CLIP-GCD~\cite{clipgcd}, which uses CLIP as the backbone, the others utilize the DINO as the backbone. To better ensure the fairness of our comparison, we also reproduce SimGCD by using CLIP as the backbone. Clearly, our TextGCD outperforms all compared methods in terms of All accuracy on all datasets. Importantly, compared to the best competitor, SimGCD, our method achieves a significant improvement on the ImageNet-1K and all the fine-grained datasets. Specifically, our method outperforms SimGCD by 7.7\% in All accuracy and 11.4\% in New accuracy on the ImageNet-1K dataset, and, by 11.1\% in All accuracy and 12.7\% in New accuracy averaged on fine-grained datasets. This demonstrates the superiority of our method over existing visual-based methods. In comparison to CLIP-GCD, which merely concatenates features from both modalities, TextGCD achieves higher results. For instance, TextGCD surpasses CLIP-GCD by 13.8\%, 16.3\%, and 10.9\% in terms of All accuracy on the CUB, Stanford Cars, and Flowers102 datasets, respectively.
\begin{table*}[!htbp]
  \caption{Results on generic datasets. The best results are highlighted in \textbf{bold}.}
  \centering
  \scriptsize
  \setlength{\tabcolsep}{1.5pt}
  \renewcommand{\arraystretch}{0.85}
  \begin{tabular*}{\linewidth}{l|@{\extracolsep{\fill}}c|ccc|ccc|ccc|ccc}
    \toprule
    \multirow{2}{*}{Methods} & \multirow{2}{*}{Backbone} & \multicolumn{3}{c|}{CIFAR-10} & \multicolumn{3}{c|}{CIFAR-100} & \multicolumn{3}{c|}{ImageNet-100} & \multicolumn{3}{c}{ImageNet-1K} \\
    \cmidrule(lr){3-5} \cmidrule(lr){6-8} \cmidrule(lr){9-11} \cmidrule(l){12-14}
    & & All & Old & New & All & Old & New & All & Old & New & All & Old & New \\
    \midrule

    GCD & DINO & 91.5 & 97.9 & 88.2 & 73.0 & 76.2 & 66.5 & 74.1 & 89.8 & 66.3 & 52.5 & 72.5 & 42.2  \\
    SimGCD & DINO & 97.1 & 95.1 & 98.1 & 80.1 & 81.2 & 77.8 & 83.0 & 93.1 & 77.9 & 57.1 & 77.3 & 46.9  \\
    DCCL & DINO & 96.3 & 96.5 & 96.9 & 75.3 & 76.8 & 70.2 & 80.5 & 90.5 & 76.2 & - & - & -  \\
    GPC & DINO & 92.2 & \textbf{98.2} & 89.1 & 77.9 & 85.0 & 63.0 & 76.9 & 94.3 & 71.0& - & -& - \\
    PromptCAL & DINO & 97.9 & 96.6 & 98.5 & 81.2 & 84.2 &75.3 & 83.1 & 92.7 & 78.3 & - & - & - \\
    SPTNet & DINO & 97.3 & 95.0 & \textbf{98.6} & 81.3 & 84.3 &75.6 & 85.4 & 93.2 & 81.4 & - & - & - \\
    
    \midrule
    CLIP-GCD & CLIP & 96.6 & 97.2 & 96.4 & 85.2 & 85.0 & \textbf{85.6} & 84.0 & \textbf{95.5} & 78.2 & - & - & -  \\
    SimGCD & CLIP & 96.6 & 94.7 & 97.5 & 81.6 & 82.6 & 79.5 & 86.1 & 94.5 & 81.9 & 48.2 & 72.7 & 36.0  \\
    TextGCD& CLIP & \textbf{98.2} & 98.0 & \textbf{98.6} & \textbf{85.7} & \textbf{86.3} & 84.6 & \textbf{88.0} & 92.4 & \textbf{85.2} & \textbf{64.8} & \textbf{77.8} & \textbf{58.3}  \\
    \bottomrule
  \end{tabular*}
  \label{tab:comparison_generic}
\end{table*}

\begin{table*}[!htbp]
  \caption{Results on fine-grained datasets. The best results are highlighted in \textbf{bold}.}
  \centering
  \scriptsize
  \setlength{\tabcolsep}{1.5pt}
  \renewcommand{\arraystretch}{0.9}
  \begin{tabular*}{\linewidth}{l|@{\extracolsep{\fill}}c|ccc|ccc|ccc|ccc}
    \toprule
    \multirow{2}{*}{Methods} & \multirow{2}{*}{Backbone} & \multicolumn{3}{c|}{CUB} & \multicolumn{3}{c|}{Stanford Cars} & \multicolumn{3}{c|}{Oxford Pets} & \multicolumn{3}{c}{Flowers102} \\
    \cmidrule(lr){3-5} \cmidrule(lr){6-8} \cmidrule(lr){9-11} \cmidrule(l){12-14}
    & & All & Old & New & All & Old & New & All & Old & New & All & Old & New \\
    \midrule

    GCD & DINO & 51.3 & 56.6 & 48.7 & 39.0 & 57.6 & 29.9 & 80.2 & 85.1 & 77.6  & 74.4 & 74.9 & 74.1 \\
    SimGCD & DINO & 60.3 & 65.6 & 57.7 & 53.8 & 71.9 & 45.0 & 87.7 & 85.9 & 88.6  & 71.3 & 80.9 & 66.5 \\
    DCCL & DINO & 63.5 & 60.8 & 64.9 & 43.1 & 55.7 & 36.2 & 88.1 & 88.2 & 88.0  & - & - & - \\
    SPTNet & DINO & 65.8 & 68.8 & 65.1 & 59.0 & 79.2 & 49.3 & - & - & -  & - & - & - \\
    \midrule
    CLIP-GCD & CLIP & 62.8 & 77.1 & 55.7 & 70.6 & \textbf{88.2} & 62.2 & - & - & -  & 76.3 & 88.6 & 70.2 \\
    SimGCD & CLIP &62.0 & 76.8 & 54.6 & 75.9 & 81.4 & 73.1 & 88.6 & 75.2 & 95.7 & 75.3 & 87.8 & 69.0\\
    TextGCD & CLIP & \textbf{76.6} & \textbf{80.6} & \textbf{74.7} & \textbf{86.9} & 87.4 & \textbf{86.7} & \textbf{95.5} & \textbf{93.9} & \textbf{96.4}  & \textbf{87.2} & \textbf{90.7} & \textbf{85.4} \\
    \bottomrule
  \end{tabular*}
  \label{tab:comparison_finegtained}
\end{table*}

\subsection{Ablation Study}
\label{abltion_studay}
\noindent \textbf{Components Ablation.} 
In Tab.~\ref{tab:ablation_components}, we present an ablation study of the key components in TextGCD, specifically cross-modality co-teaching, contrastive learning, and soft voting. Beginning with the Baseline method, we incrementally incorporate these three components into the framework to evaluate their impact on the performance of both image and text classifiers. We also report the results of 1) employing the first tag derived from the visual lexicon as the GCD labels and 2) performing $k$-means~\cite{kmeans} using the image or text features without training. It becomes evident that the ``First-Tag'' and ``$k$-means'' methods yield inferior results compared to the baseline. On the other hand, the ``$k$-means'' method and the baseline results demonstrate that using the textual information achieves clearly higher results than using the image cues, thereby highlighting the superior quality of the generated textual descriptions. Compared to the baseline, integrating our proposed co-teaching approach markedly elevates the performance across both modalities, particularly notable within the context of the fine-grained CUB dataset. This substantiates the efficacy of our co-teaching strategy in fostering synergistic learning and collaborative advancement between text and image models. Furthermore, image-text contrastive learning consistently increases the All and New accuracies, underlining the necessity of aligning the two modalities. Soft voting can further improve performance, illustrating that the integration of information from both modalities leads to more precise category discrimination.

\begin{table}[!t]
  \caption{Ablation study on TextGCD components. ``T'' and ``I'' denote the output results of the text and image classifiers, respectively. ``First-Tag'' refers to the accuracy achieved using the most similar tag from the visual lexicon as the label. ``$k$-means'' indicates applying $k$-means on the initial backbone features.}
  \centering
  \scriptsize
  \setlength{\tabcolsep}{3pt}
  \renewcommand{\arraystretch}{1.0}
  \begin{tabular}{lc|ccc|ccc}
    \toprule
    \multirow{2}{*}{Methods} & \multirow{2}{*}{Classifier} & \multicolumn{3}{c|}{CIFAR-100} & \multicolumn{3}{c}{CUB} \\
    \cmidrule(lr){3-5} \cmidrule(lr){6-8}
    & & All & Old & New & All & Old & New \\
    \midrule
    First-Tag & - & 12.0 & 12.0 & 11.9 & 47.8 & 43.4 & 52.1\\
    \midrule
    \( k \)-means & T & 70.2 & 66.6 & 77.5 & 67.6 & 64.5 & 69.2 \\
    & I & 46.5 & 45.6 & 48.5 & 46.7 & 50.6 & 44.7 \\
    \hline
    Baseline & T & 81.8 & 83.6 & 78.3 & 67.0 & 75.9 & 62.6 \\
    & I & 76.3 & 80.9 & 67.0 & 49.0 & 64.2 & 41.3 \\
    \hline
    \( + \text{Co-Teaching} \) & T & 84.4 & 86.2 & 81.1 & 74.4 & 79.3 & 72.0 \\
    & I & 82.4 & 85.0 & 77.3 & 71.7 & 79.4 & 67.8 \\
    \hline
    \( + \text{Contrastive} \) & T & 84.6 & 84.8 & 84.1 & 74.8 & 78.6 & 73.0 \\
    & I & 83.1 & 84.0 & 81.1 & 73.9 & 80.4 & 70.7 \\
    \hline
    \( + \text{Soft Voting} \) & - & \textbf{85.7} & \textbf{86.3} & \textbf{84.6} & \textbf{76.6} & \textbf{80.6} & \textbf{74.7} \\
    \bottomrule
  \end{tabular}
  \label{tab:ablation_components}
\end{table}

\begin{table}[h]
  \caption{Ablation study on TextGCD training stages. ``Warm-up'' refers to the warm-up stage and ``Cls. Align'' denotes the class-aligning stage. ``T$\to$I'' signifies aligning the image model using the text model, whereas ``I$\to$T'' indicates aligning the text model using the image model.}
  \centering
  \scriptsize
  \setlength{\tabcolsep}{2.0pt}
  \renewcommand{\arraystretch}{0.95}
  \begin{tabular}{cccc|ccc|ccc|ccc|ccc}
    \toprule
   Warm- & \multicolumn{2}{c}{Cls. Align} & Co- & \multicolumn{3}{c|}{CIFAR-100} & \multicolumn{3}{c|}{ImageNet-100} & \multicolumn{3}{c|}{CUB} & \multicolumn{3}{c}{Stanford Cars} \\
    \cmidrule(lr){2-3} \cmidrule(lr){5-7} \cmidrule(lr){8-10} \cmidrule(lr){11-13} \cmidrule(lr){14-16}
    up& I$\to$T & T$\to$I & Teaching& All & Old & New & All & Old & New & All & Old & New & All & Old & New \\
    \midrule
    & & & \checkmark &  83.3 & 86.8 & 76.4 & 80.6 & 92.8 & 74.5 & 74.8 & 77.4 & 73.5 & 83.9 & 88.1 & 80.6 \\
    \checkmark&  & & \checkmark& 82.8 & \textbf{87.2} & 74.1 & 81.6 & 92.6 & 76.0 & 76.5 & \textbf{81.2} & 74.1 & 85.3 & \textbf{87.8} & 84.0 \\
    \checkmark & \checkmark &  & \checkmark & 84.6 & 86.6 & 80.6 & 87.5 & \textbf{93.5} & 84.4 & 76.1 & 80.9 & 73.6 & \textbf{87.5} & 86.4 & \textbf{87.9} \\
    \checkmark &  & \checkmark & \checkmark & \textbf{85.7} & 86.3 & \textbf{84.6} & \textbf{88.0} & 92.4 & \textbf{85.2} & \textbf{76.6} & 80.6 & \textbf{74.7} & 86.9 & 87.4 & 86.7 \\

    \bottomrule
  \end{tabular}
  \label{tab:ablation_stages}
\end{table}
\noindent \textbf{Training Stages Ablation.} 
In Tab.~\ref{tab:ablation_stages}, we assess the impact of each stage within the proposed CCT phase, \textit{i.e.,} warm-up, class-aligning, and co-teaching. Results show that using co-teaching solely achieves lower results than our full method. Specifically, the warm-up process is important for the fine-grained dataset, \textit{i.e.}, CUB and SCars. On the other hand, the class-aligning consistently improves the performance on all four datasets, which obtains 10.5\%, 9.2\%, 0.6\%, and 2.7\% in New accuracy on CIFAR-100, ImageNet-100, CUB and SCars, respectively. This observation highlights the vital role of class-aligning stage in resolving conflicts between the two classifiers. 
To further investigate the proposed class-aligning strategy, we evaluate the variant of using the image model to guide the text model, denoted as ``I$\to$T''.
In terms of All and New accuracies, we observe that 1) both strategies can achieve consistent improvements (except for ``I$\to$T'' on CUB) and that 2) the proposed text-guide-image strategy produces higher results than the image-guide-text variant in most cases. These advantages can be attributed to the more representative initial features generated by the textual descriptions. Therefore, the discriminative textual features can generate high-quality pseudo-labels for effectively guiding visual model (text-guide-image) or can be robust to noisy labels when guided by the visual model (image-guide-text). This further demonstrates the importance of introducing textual modality into our co-teaching framework.

\subsection{Evaluation}
\label{sec:evaluation}
\noindent \textbf{Backbone Evaluation.} In Tab.~\ref{tab:evaluation_backbone}, we evaluate the impact of the image encoder in TextGCD by substituting the image encoder of CLIP with that of DINO, while maintaining the text encoder from CLIP. In TextGCD(DINO), we omit the cross-modal contrastive loss $\mathcal{L}_{\text{con}}$. Despite a decreased accuracy on CUB, TextGCD(DINO) still significantly outperforms SimGCD, which utilizes the same image encoder.
\begin{table}[th]
  \caption{Evaluation on different pretrained ViT-B backbones.}
  \centering
  \scriptsize
  \setlength{\tabcolsep}{4pt}
  \renewcommand{\arraystretch}{0.85}
  \begin{tabular}{l c cccc ccc}
    \toprule
    \multirow{2}{*}{Methods} & \multirow{2}{*}{Backbone} & \multicolumn{3}{c}{CIFAR-100} & \multicolumn{3}{c}{CUB} \\
    \cmidrule(lr){3-5} \cmidrule(lr){6-8}
    & & All & Old & New & All & Old & New \\
    \midrule
    SimGCD & DINO & 80.1 & 81.2 & 77.8 & 60.3 & 65.6 & 57.7 \\
    TextGCD & DINO & \textbf{86.1} & \textbf{88.7} & 81.0 & 73.7 & 80.3 & 70.4 \\
    \midrule
    TextGCD & CLIP & 85.7 & 86.3 & \textbf{84.6} & \textbf{76.6} & \textbf{80.6} & \textbf{74.7} \\
    \bottomrule
  \end{tabular}
  \label{tab:evaluation_backbone}
\end{table}
Moreover, we compare SimGCD and TextGCD using the ViT-H-based CLIP for a more fair comparison, as we use the ViT-H for text generation. Tab.~\ref{tab:evaluation_backbone_vith} shows that our TextGCD outperforms SimGCD by a large margin on both datasets. Interestingly, both SimGCD and our method show a decrease in performance on the generic dataset CIFAR-100, in terms of New accuracy, compared to the ones using ViT-B-16 as the backbone (see Tab.~\ref{tab:comparison_generic}). This suggests that fine-tuning on the generic dataset may adversely affect the ability of the large model. In addition, we can observe a small margin between our methods with ViT-B-16 and ViT-H-14. This indicates that our advantage mainly benefited from the high-quality text description and the co-teaching strategy, enabling our approach to adapt to both smaller and larger models.
\begin{table}[th]
  \caption{Evaluation on ViT-H-14 backbone.}
  \centering
  \scriptsize
  \setlength{\tabcolsep}{3pt}
  \renewcommand{\arraystretch}{0.85}
  \begin{tabular}{l c cccc ccc}
    \toprule
    \multirow{2}{*}{Methods} & \multirow{2}{*}{CLIP Backbone} & \multicolumn{3}{c}{CIFAR-100} & \multicolumn{3}{c}{CUB} \\
    \cmidrule(lr){3-5} \cmidrule(lr){6-8}
    & & All & Old & New & All & Old & New \\
    \midrule
    SimGCD &ViT-H-14 & 78.1 & 80.0 & 74.4 & 69.1 & 76.3 & 65.4 \\
    TextGCD &ViT-H-14 & \textbf{86.4} & \textbf{89.3} & \textbf{80.7} & \textbf{78.6} & \textbf{81.5} & \textbf{77.1} \\
    \bottomrule
  \end{tabular}
  \label{tab:evaluation_backbone_vith}
\end{table}

\noindent \textbf{Text Generation Evaluation.} 
In Tab.~\ref{tab:evalutation_text}, we analyze the impact of the number of tags and attributes for constructing category descriptions in the RTG phase. Results show that richer textual content (more tags and attributes) leads to higher category recognition accuracy. Due to the token limitation of CLIP, we can use up to only three tags and two attributes in the input of the text model. We will handle this drawback and investigate the impact of including more tags and attributes in future work.
\begin{table}[!th]
  \caption{Evaluation on the number of tags and attributes.}
  \centering
  \scriptsize
  \setlength{\tabcolsep}{4.8pt}
  \renewcommand{\arraystretch}{0.85}
  \begin{tabular}{cc cccc ccc}
    \toprule
    \multirow{2}{*}{\# Tag} & \multirow{2}{*}{\# Attribute} & \multicolumn{3}{c}{CIFAR-100} & \multicolumn{3}{c}{CUB} \\
    \cmidrule(lr){3-5} \cmidrule(lr){6-8}
    & & All & Old & New & All & Old & New \\
    \midrule
    1 & 0 & 81.6 & 81.6 & 81.8 & 67.9 & 69.6 & 67.0 \\
    2 & 0 & 83.5 & 83.4 & 83.7 & 68.6 & 75.3 & 65.3 \\
    3 & 0 & 83.7 & 84.7 & 81.6 & 69.9 & 74.9 & 67.3 \\
    3 & 1 & 84.7 & 85.4 & 83.4 & 74.7 & 80.5 & 71.7 \\
    3 & 2 & \textbf{85.7} & \textbf{86.3} & \textbf{84.6} & \textbf{76.6} & \textbf{80.6} & \textbf{74.7} \\
    \bottomrule
  \end{tabular}
  \label{tab:evalutation_text}
\end{table}

\noindent \textbf{Auxiliary Model Evaluation.} 
In Tab.~\ref{tab:evaluation_auxiliarymodel}, we explore the effect of substituting the auxiliary model with FLIP~\cite{flip} and CoCa~\cite{coca}. Results show that our method consistently produces high performance on both datasets and that using a more powerful model (FLIP) leads to higher results.
\begin{table}[h]
  \caption{Evaluation on the auxiliary model.}
  \centering
  \scriptsize
  \setlength{\tabcolsep}{3.75pt}
  \renewcommand{\arraystretch}{0.85}
  \begin{tabular}{lccccccc}
    \toprule
    Auxiliary& \multirow{2}{*}{Backbone} & \multicolumn{3}{c}{CIFAR-100} & \multicolumn{3}{c}{CUB} \\
    \cmidrule(lr){3-5} \cmidrule(lr){6-8}
    Model & & All & Old & New & All & Old & New \\
    \midrule
    CLIP &ViT-H-14  & 85.7 & 86.3 & 84.6 & 76.6 & 80.6 & 74.7 \\
    CoCa & ViT-L-14  & 85.2 & 85.8 & 83.8 & 73.6 & 81.7 & 69.5 \\
    FLIP & ViT-G-14 & \textbf{87.6} & \textbf{87.7} & \textbf{87.5} & \textbf{79.5} & \textbf{83.5} & \textbf{77.4} \\
    \bottomrule
  \end{tabular}
  \label{tab:evaluation_auxiliarymodel}
\end{table}

\noindent \textbf{Co-Teaching Strategy Evaluation.} 
To demonstrate the effectiveness of the cross-modal co-teaching strategy, we conduct a comparative analysis with a co-teaching approach utilizing a single modality. Results in Tab.~\ref{tab:evaluation_coteaching} show that single-modality co-teaching variants obtain lower results than the proposed cross-modality co-teaching. This finding highlights the critical role of ensuring the model diversity in co-teaching.
\begin{table}[th]
  \caption{Evaluation on co-teaching strategies. ``I$\leftrightarrow$T'' is our cross-modality co-teaching. ``T$\leftrightarrow$T'' and ``I$\leftrightarrow$I'' indicate the single-modality variant using text modality and image modality, respectively.}
  \centering
  \scriptsize
  \setlength{\tabcolsep}{6.5pt}
  \renewcommand{\arraystretch}{0.85}
  \begin{tabular}{lcccccc}
    \toprule
    \multirow{2}{*}{Methods} & \multicolumn{3}{c}{CIFAR-100} & \multicolumn{3}{c}{CUB} \\
    \cmidrule(lr){2-4} \cmidrule(lr){5-7}
    & All & Old & New & All & Old & New \\
    \midrule
    T$\leftrightarrow$T & 79.8 & 83.6 & 72.1 & 71.7 & 73.0 & 71.1 \\ 
    I$\leftrightarrow$I & 75.5 & 80.5 & 65.7 & 60.3 & 79.5 & 50.7 \\ 
    I$\leftrightarrow$T (Ours) & \textbf{85.7} & \textbf{86.3} & \textbf{84.6} & \textbf{76.6} & \textbf{80.6} & \textbf{74.7} \\
    \bottomrule
  \end{tabular}
  \label{tab:evaluation_coteaching}
\end{table}

\subsection{Hyper-parameters Analysis}
\label{sec:hyperparameters}

\begin{figure}[h]
  \centering
    \includegraphics[width=0.9\linewidth]{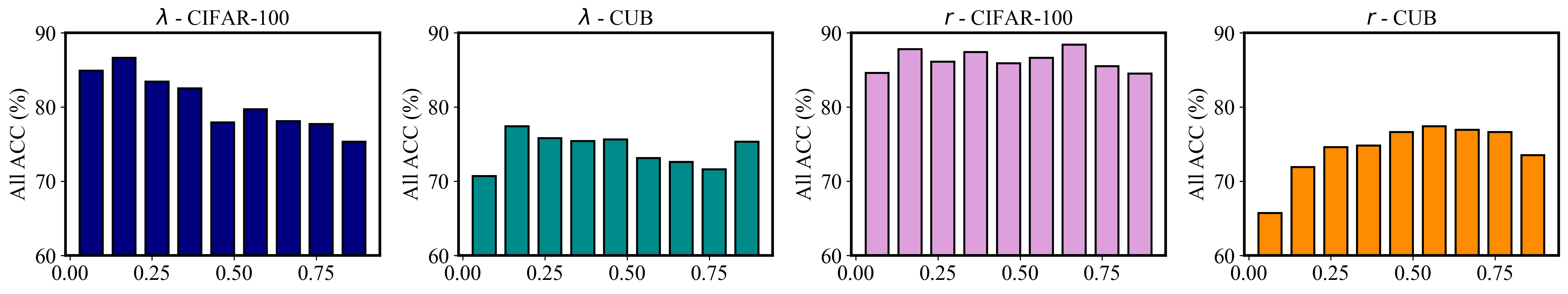}

   \caption{Impact of hyper-parameters. 
   }
   \label{fig:hyperparameters}
\end{figure}

\noindent \textbf{Balancing Factor $\lambda$ and  Proportion Coefficient $r$.} The impact of the balancing factor $\lambda$ and the proportion coefficient $r$ is illustrated in Fig.~\ref{fig:hyperparameters}. We use All accuracy as the evaluation metric. An increase in $\lambda$ suggests that the classifiers of both models increasingly prioritize labeled data from known categories. Results show that $\lambda=0.2$ achieves the best accuracy on both datasets. The parameter $r$ determines the fraction of high-confidence samples chosen from each category for the text and image models. A higher value of $r$ is typically advantageous, and the best result is achieved when $r=0.6$. Note that we use the same hyper-parameters ($\lambda$ and $r$) for all datasets to avoid over-tuning.

\section{Conclusion}
\label{sec:conclusion}
In this paper, we introduce a novel approach called TextGCD for Generalized Category Discovery. Compared to previous methods that only rely on visual patterns, we additionally integrate textual information into the framework. Specifically, we first construct a visual lexicon based on off-the-shelf category tags and enrich it with attributes by Large Language Models (LLMs). Following this, a Retrieval-based Text Generation (RTG) approach is proposed to assign each sample with likely tags and corresponding attributes. Building upon this, we design a Cross-modal Co-Teaching (CCT) strategy to fully take the mutual benefit between visual and textual information, enabling robust and complementary model training. Experiments on eight benchmarks show that our TextGCD produces new state-of-the-art performance. We hope this study could bring a new perspective, which assists the framework with foundational model and textual information, for the GCD community.

\section*{Acknowledgement}
We acknowledge the support of the MUR PNRR project iNEST-Interconnected Nord-Est Innovation Ecosystem (ECS00000043) funded by the NextGenerationEU. Also, this work was partially supported by the EU Horizon projects ELIAS (No. 101120237) and AI4Trust (No. 101070190).


%
%
\bibliographystyle{splncs04}
\bibliography{main}

\clearpage
\newpage

\input{supplement}

\end{document}

%% file: supplement.tex
\title{Supplementary Material}

\titlerunning{Cross-Modality Co-Teaching for
Generalized Visual Class Discovery}

\author{Haiyang Zheng\inst{1}\textsuperscript{$\star$}
\and
Nan Pu\inst{1}\textsuperscript{$\star$}
\and
Wenjing Li\inst{2,3}\textsuperscript{$\dagger$}
 \and
Nicu Sebe\inst{1}
\and
Zhun Zhong\inst{2,3}\textsuperscript{$\dagger$}
}

\authorrunning{H. Zheng et al.}

\institute{\textsuperscript{1}University of Trento \textsuperscript{2}Hefei University of Technology
\textsuperscript{3}University of Nottingham
\blankfootnote{\textsuperscript{$\star$} Equal contribution. \quad \textsuperscript{$\dagger$} Corresponding author.}
}

\maketitle

\section*{A. Datasets Settings}

In the main paper, we follow the GCD protocol~\cite{gcd} to determine the number of known classes in the dataset and the number of labeled data samples selected from these classes. We conduct extensive experiments across four generic image classification datasets, namely, CIFAR-10, CIFAR-100~\cite{cifar}, ImageNet-100, and ImageNet-1K~\cite{imagenet}, and four fine-grained image classification datasets, specifically, CUB~\cite{cub}, Stanford Cars~\cite{scars}, Oxford Pets~\cite{pets}, and Flowers102~\cite{flowers}. Detailed dataset information is provided in Table~\ref{tab:datasets}.

\begin{table}[!th]
  \centering
  \caption{Statistics of the datasets for GCD.}
  \scriptsize
  \renewcommand{\arraystretch}{0.85}
\begin{tabular}{@{}lcccc@{}}
    \toprule
    \multirow{2}{*}{\textbf{Dataset}} & \multicolumn{2}{c}{Labelled} & \multicolumn{2}{c}{Unlabelled} \\

    \cmidrule(lr){2-3} \cmidrule(lr){4-5} 
    &\textbf{\#Image} & \textbf{\#Class} & \textbf{\#Image} & \textbf{\#Class} \\
    \midrule
    ImageNet-100~\cite{imagenet} & 31.9K & 50 & 95.3K & 100 \\
    ImageNet-1K~\cite{imagenet} & 321K & 500 & 960K & 1000 \\
    CIFAR-10~\cite{cifar} & 12.5K & 5 & 37.5K & 10 \\
    CIFAR-100~\cite{cifar} & 20.0K & 80 & 30.0K & 100 \\
    CUB~\cite{cub} & 1.5K & 100 & 4.5K & 200 \\
    Stanford Cars~\cite{scars} & 2.0K & 98 & 6.1K & 196 \\
    Oxford Pets~\cite{pets} & 0.9K & 19 & 2.7K & 37 \\
    Flowers102~\cite{flowers} & 0.3K & 51 & 0.8K & 102 \\
    \bottomrule
  \end{tabular}
  \label{tab:datasets}
\end{table}

\section*{B. Extended Experiment Results}

\subsection*{B.1. Main Results}
\noindent \textbf{Mean \& Std.} In the main paper, we present the full results as the average of three runs to mitigate the impact of randomness. Detailed outcomes for TextGCD, encompassing mean values and population standard deviation, are delineated in Tab.~\ref{tab:complete_results}.
\begin{table}[th]
  \caption{Mean and Std of Accuracy in Three Independent Runs}
  \centering
  \scriptsize
  \setlength{\tabcolsep}{8pt}
  \renewcommand{\arraystretch}{1}
  \begin{tabular}{lccc}
    \toprule
    Dataset & All & Old & New \\
    \midrule
    CIFAR10 \cite{cifar} & 98.2$\pm$0.3 & 98.0$\pm$0.1 & 98.6$\pm$0.5 \\
    CIFAR100 \cite{cifar} & 85.7$\pm$0.6 & 86.3$\pm$0.6 & 84.6$\pm$0.9 \\
    ImageNet-100 \cite{imagenet} & 88.0$\pm$0.7 & 92.4$\pm$0.3 & 85.2$\pm$1.2 \\
    ImageNet-1K \cite{imagenet} & 64.8$\pm$0.1 & 77.8$\pm$0.1 & 58.3$\pm$0.1 \\
    CUB \cite{cub} & 76.6$\pm$1.0 & 80.6$\pm$0.8 & 74.7$\pm$1.1 \\
    Stanford Cars \cite{scars} & 86.9$\pm$0.6 & 87.4$\pm$1.0 & 86.7$\pm$0.4 \\
    Oxford Pets \cite{pets} & 95.5$\pm$0.1 & 93.9$\pm$0.6 & 96.4$\pm$0.1 \\
    Flowers102 \cite{flowers} & 87.2$\pm$1.7 & 90.7$\pm$0.9 & 85.4$\pm$2.5 \\
    \bottomrule
  \end{tabular}
  \label{tab:complete_results}
\end{table}

\begin{figure}[th]
  \centering
  \begin{subfigure}[b]{0.48\linewidth}
    \includegraphics[width=\linewidth]{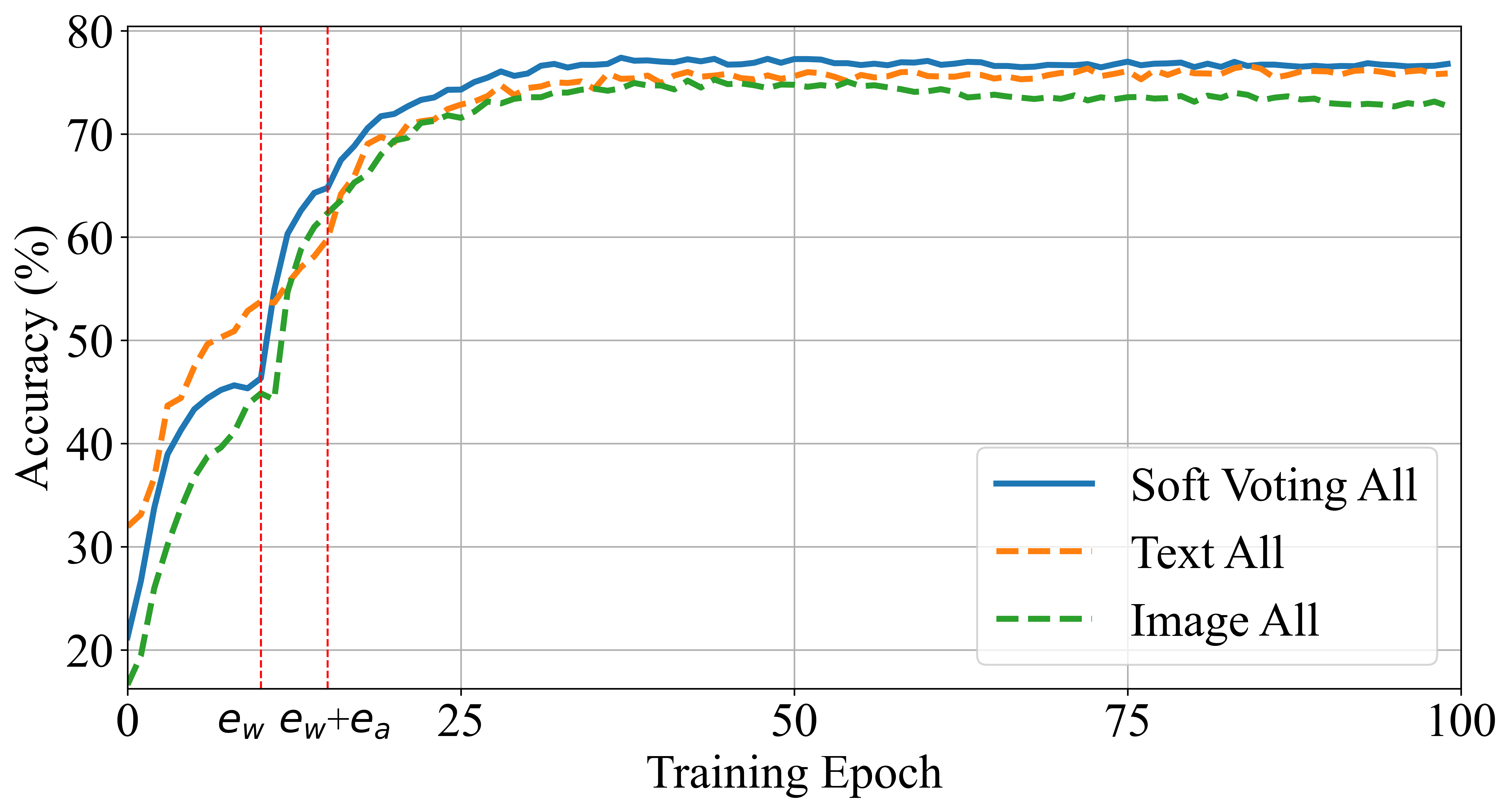}
    \caption{All accuracy evolution across epochs.}
    \label{fig:all_hyperparameters}
  \end{subfigure}
  \hfill 
  \begin{subfigure}[b]{0.48\linewidth}
    \includegraphics[width=\linewidth]{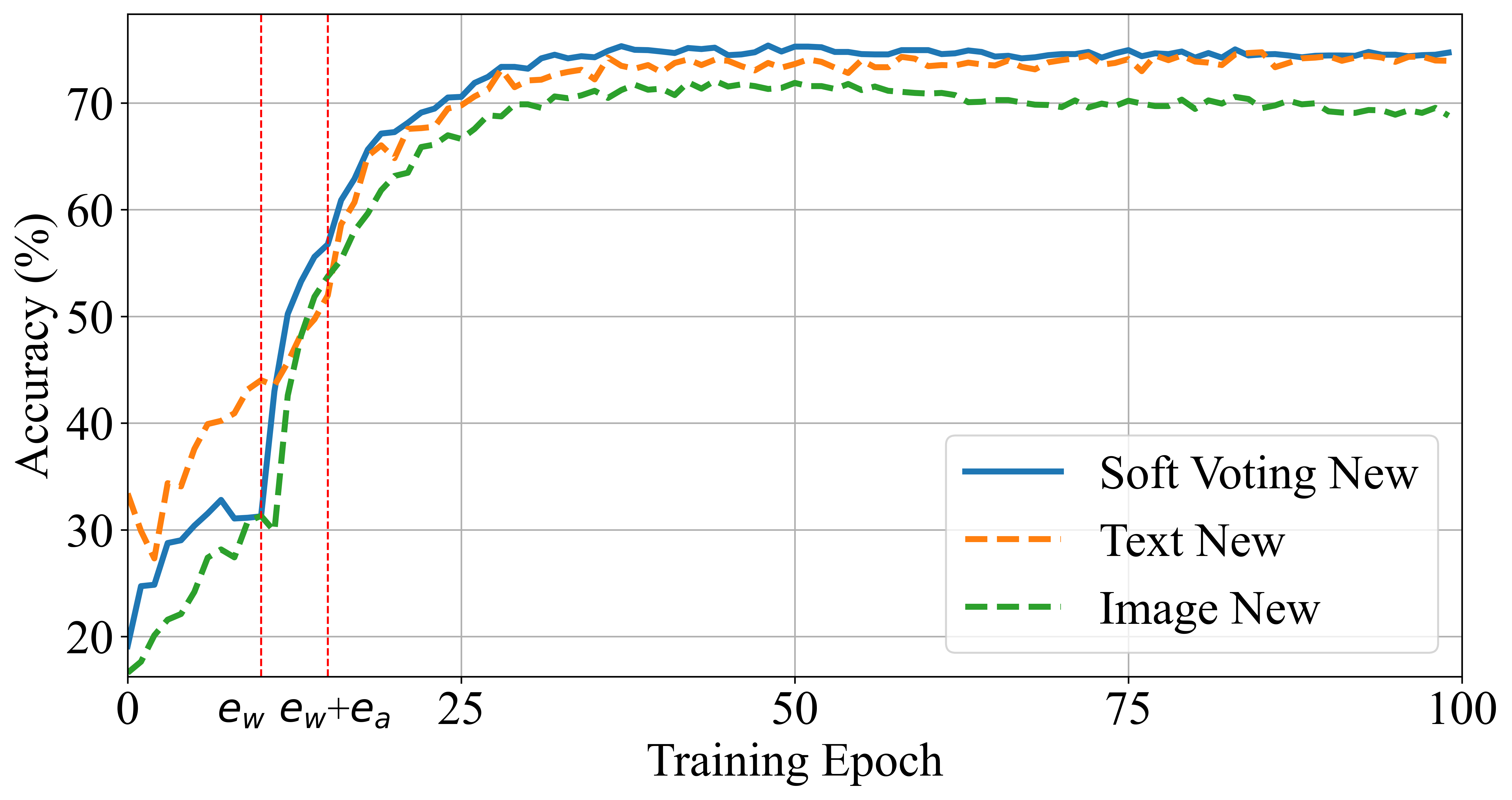}
    \caption{New accuracy evolution across epochs.}
    \label{fig:new_hyperparameters}
  \end{subfigure}
  
  \caption{Training accuracy over epochs on the CUB dataset.}
  \label{fig:results_over_epochs}
\end{figure}

\noindent \textbf{Results Through Training.} We employ the CUB dataset as a case study to illustrate the model's predictive performance evolution during training, depicted as a curve across epochs. This representation includes results from the text model, the image model, and the soft voting mechanism, as indicated in Fig.~\ref{fig:results_over_epochs}. The graph reveals two key observations: 1) Implementing the class-aligning stage (commencing at the $e_w$-th epoch) and the co-teaching stage (initiating at the $e_w$+$e_a$-th epoch) markedly improves the identification of previously unseen categories. 2) Integrating insights from different modalities via soft voting enhances the overall prediction accuracy.

\subsection*{B.2. Additional Baseline Results}
In Tab.~\ref{tab:two_baseline}, we present a comparison between the results of two baselines and our TextGCD. The term ``Zero-shot'' refers to the zero-shot results obtained using CLIP~\cite{clip} in the GCD setting, given the names of all categories (including new categories). ``Frozen-text'' refers to partially fine-tuning the image encoder only while using frozen text embeddings of the category labels as classifiers. This approach treats the problem as partially supervised, where only some classes are unannotated, but all class names are known. As shown in Tab.~\ref{tab:two_baseline}, our TextGCD significantly outperforms these two baselines.
\begin{table}[th]
  \centering
  \scriptsize
\caption{Comparison of results between two baselines with known text descriptors and our TextGCD.}
  \begin{tabular}{l|ccccccccc}
    \toprule
      \multirow{2}{*}{Methods}& \multicolumn{3}{c}{CIFAR100} & \multicolumn{3}{c}{CUB} & \multicolumn{3}{c}{Stanford Cars} \\ 
     \cmidrule(lr){2-4} \cmidrule(lr){5-7} \cmidrule(lr){8-10}
      & All  & Old  & New & All  & Old  & New & All  & Old  & New \\
    \midrule
    Zero-shot & 66.2  &  64.6 & 69.5  & 54.3 & 62.3 & 50.3 & 67.3 & 59.4 & 71.3 \\ 
    Frozen-text & 79.2  &  81.5 & 74.6  & 62.6 & 80.2 & 53.9 & 74.9 & 77.9 & 73.4 \\ 
    TextGCD & \textbf{85.7}  &  \textbf{86.3} & \textbf{84.6}  & \textbf{76.6} & \textbf{80.6} & \textbf{74.7} & \textbf{86.9} & \textbf{87.4} & \textbf{86.7} \\ 
    \bottomrule
  \end{tabular}
  \label{tab:two_baseline}
\end{table}

\subsection*{B.3. Evaluation on Varying Known Knowledge}

\begin{figure}[t]
  \centering
    \includegraphics[width=0.8\linewidth]{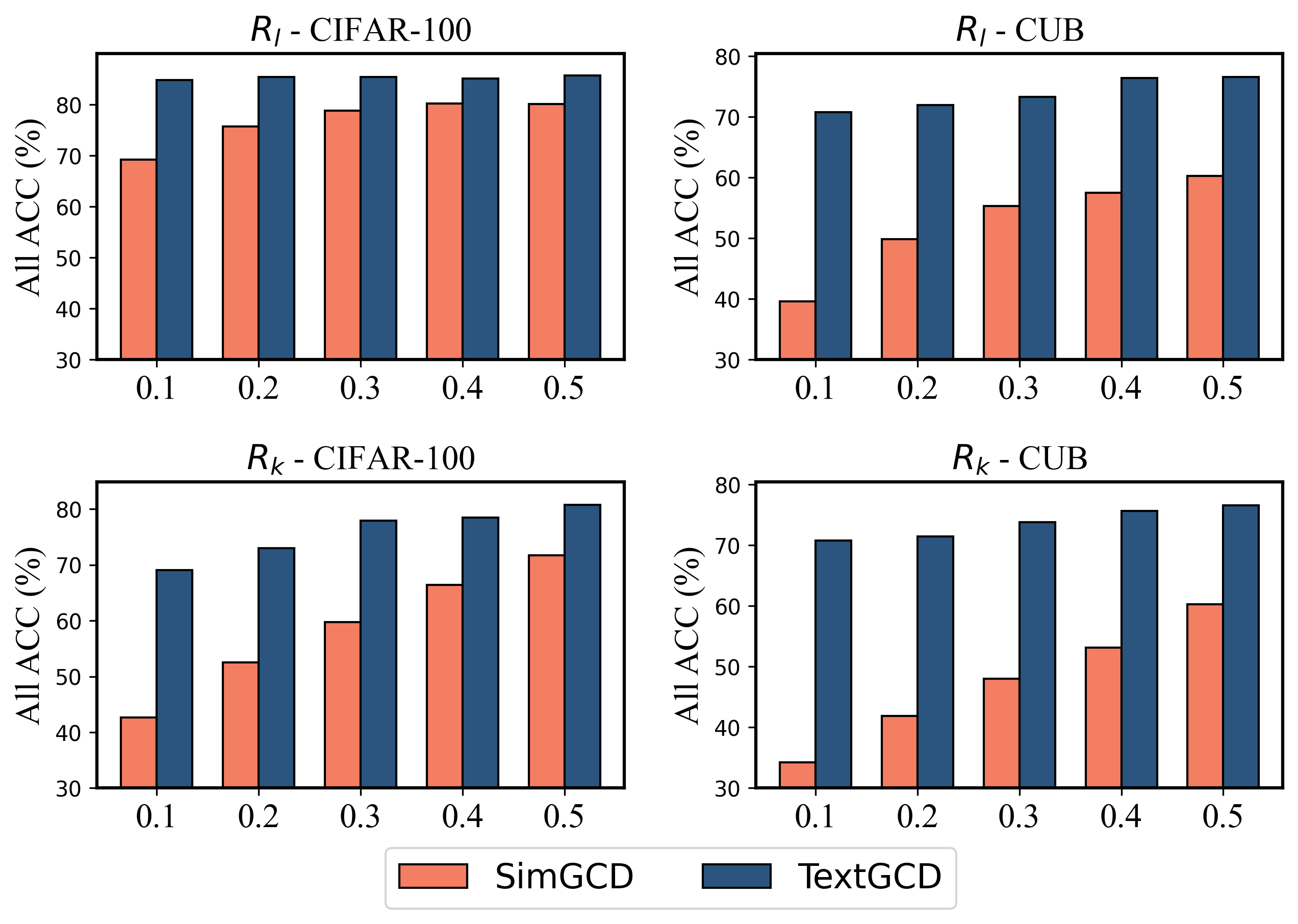}

   \caption{Comparative analysis of TextGCD and SimGCD with varying hyper-parameters $R_k$ and $R_l$.}
   \label{fig:less_prior}
\end{figure}

In the main paper, we adhere to the GCD protocol~\cite{gcd} for setting the number of known classes in the dataset and the number of labeled data samples selected from these classes. The number of selected known categories is denoted as $K_L=R_k * K$. The variable $R_l$ represents the proportion of data labeled from each known category. Therefore, $R_k$ and $R_l$ together determine the amount of labeled data. Following~\cite{gcd}, we set $R_k$ to 0.5 for most datasets, except for CIFAR-100, where $R_k$ is set to 0.8. $R_l$ is maintained at 0.5 across all datasets. We compare SimGCD~\cite{simgcd} with our TextGCD on the CIFAR-100 and CUB datasets to assess the influence of hyper-parameters $R_k$ and $R_l$. To minimize the effects of randomness, each experiment is conducted three times, and the average All accuracy is reported.

Lower values of $R_k$ and $R_l$ correspond to reduced prior knowledge, thereby elevating the difficulty level of the GCD task. As shown in Fig.~\ref{fig:less_prior}, in scenarios characterized by limited prior knowledge, our TextGCD method demonstrates a notable advantage over SimGCD. Notably, when merely 10\% of data from each known category is designated as labeled data, TextGCD exhibits an improvement of 31.2\% over SimGCD.  This exceptional performance can be attributed to two main reasons: firstly, the introduction of the textual modality brings additional prior knowledge; secondly, the accurate prior knowledge of the textual modality is obtained during the RTG phase.

\subsection*{B.4. Additional Results with Unknown $K$}
In the main paper, we follow the common practice \cite{simgcd} of assuming the total number of categories is known a priori. Here, we explore the scenario where such knowledge is unknown. Vaze et al. \cite{gcd} provides an off-the-shelf method to estimate the class number in unlabeled data. We introduce CLIP’s visual feature into the off-the-shelf method. As shown in Tab.~\ref{tab:unknown_nums}, we present the estimated number of categories and the error rates for CIFAR100, CUB, and Stanford Cars. With the estimation of the number of categories, we compared our results with the SimGCD algorithm on these three datasets. As shown in Tab.~\ref{tab:unknown_nums}, our TextGCD surpasses SimGCD (CLIP backbone) by an average of 9.9\% in terms of All accuracy across the three datasets.

\begin{table}[th]
  \centering
  \scriptsize
\caption{Comparison with SimGCD with unknown category numbers. The upper part of the table shows the estimated number of categories, while the lower part presents the accuracy comparison.}
  \begin{tabular}{l|ccccccccc}
    \toprule
     & \multicolumn{3}{c}{CIFAR100} & \multicolumn{3}{c}{CUB} & \multicolumn{3}{c}{Stanford Cars} \\ 
     \midrule
    Ground truth $K$ & \multicolumn{3}{c}{100} & \multicolumn{3}{c}{200} & \multicolumn{3}{c}{196} \\ 
    Estimation & \multicolumn{3}{c}{102} & \multicolumn{3}{c}{193} & \multicolumn{3}{c}{212} \\ 
    Error & \multicolumn{3}{c}{2\%} & \multicolumn{3}{c}{3.5\%} & \multicolumn{3}{c}{8.2\%} \\ 
    \midrule
      & All  & Old  & New & All  & Old  & New & All  & Old  & New \\
    \midrule
    SimGCD(DINO) & 80.6  &  81.0 & 79.1  & 62.0 & 67.1 & 59.3 & 48.6 & 64.8 & 40.7 \\ 
    SimGCD(CLIP) & 82.3  &  83.5 & 80.1  & 60.2 & 74.5 & 52.9 & 70.4 & 75.0 & 67.3 \\ 
    Ours & \textbf{86.4}  &  \textbf{87.2} & \textbf{84.7}  & \textbf{74.8} & \textbf{80.3} & \textbf{72.0} & \textbf{81.5} & \textbf{77.8} & \textbf{83.2} \\ 
    \bottomrule
  \end{tabular}
  \label{tab:unknown_nums}
\end{table}

\subsection*{B.5. Evaluation on Pseudo Label Forms}
In Tab.~\ref{tab:evaluation_pseduolabel}, we evaluate the effectiveness of different pseudo-labeling approaches in the co-teaching process. Our results demonstrate that the use of hard pseudo labels significantly surpasses soft pseudo labels, with an average improvement of 6.8\% in All accuracy across both datasets. This finding suggests that hard pseudo labels are more advantageous for the co-teaching process.
\begin{table}[!htbp]
  \centering
  \caption{Evaluation on pseudo label forms. ``Soft'' denotes the use of predicted probabilities from each model (image and text) as pseudo labels for the other, while ``Hard'' indicates the application of hard pseudo labels in co-teaching.}
  \scriptsize
  \setlength{\tabcolsep}{5pt}
  \renewcommand{\arraystretch}{0.9}
  \begin{tabular}{lcccccc}
    \toprule
    \multirow{2}{*}{Methods} & \multicolumn{3}{c}{CIFAR 100} & \multicolumn{3}{c}{CUB} \\
    \cmidrule(lr){2-4} \cmidrule(lr){5-7}
    & All & Old & New & All & Old & New \\
    \midrule
    Soft & 77.6 & 79.3 & 74.1 & 72.9 & 77.2 & 70.8 \\ 
    Hard & \textbf{85.7} & \textbf{86.3} & \textbf{84.6} & \textbf{76.6} & \textbf{80.6} & \textbf{74.7} \\
    \bottomrule
  \end{tabular}
  \label{tab:evaluation_pseduolabel}
\end{table}

\subsection*{B.6. Generating Text for Images Using BLIP}
BLIP~\cite{blip} introduces an optimization objective for generative tasks, facilitating the generation of textual descriptions for images. We explore using BLIP~\cite{blip} (blip-large) and BLIP2~\cite{blip2} (blip2-opt-2.7b) to generate textual content for images. For equitable comparison, we employ prompts such as ``The name of this object/bird is'' for tag generation and ``The feature of this object/bird is'' for attribute generation. These elements—three tags and two attributes—are then amalgamated to construct descriptive texts for images. As demonstrated in Tab.~\ref{tab:blip}, although using BLIP2 for text generation shows some improvement on the general dataset CIFAR-100, the methods employing BLIP and BLIP2 yield results significantly inferior to our retrieval-based method on the fine-grained dataset CUB. In particular, our method surpasses the BLIP2-based method by 35.7\% in All accuracy. This underperformance is attributed to BLIP's tendency to overlook detailed content, thus limiting its accuracy in fine-grained class discovery.

\begin{table}[!htbp]
  \centering
  \caption{Comparative analysis of text generation using BLIP.}
  \scriptsize
  \setlength{\tabcolsep}{4pt}
  \renewcommand{\arraystretch}{0.75}
  \begin{tabular}{lcccccc}
    \toprule
    \multirow{2}{*}{Methods} & \multicolumn{3}{c}{CIFAR 100} & \multicolumn{3}{c}{CUB} \\
    \cmidrule(lr){2-4} \cmidrule(lr){5-7}
    & All & Old & New & All & Old & New \\
    \midrule
    BLIP-based &  74.1 & 75.6  & 71.0 &  36.2 & 48.7  & 29.9  \\ 
    BLIP2-based &  \textbf{87.8} & \textbf{87.9}  &\textbf{ 87.5}  &  40.9 & 59.3  & 31.6  \\
    Retrieval-based (Ours) & 85.7  & 86.3  & 84.6  &  \textbf{76.6} & \textbf{80.6}  & \textbf{74.7}  \\
    \bottomrule
  \end{tabular}
  \label{tab:blip}
\end{table}

\subsection*{B.7. Visualization of Feature Distributions}
\begin{figure}[!htbp]
  \centering
  \begin{subfigure}{.24\linewidth}
    \includegraphics[width=\linewidth]{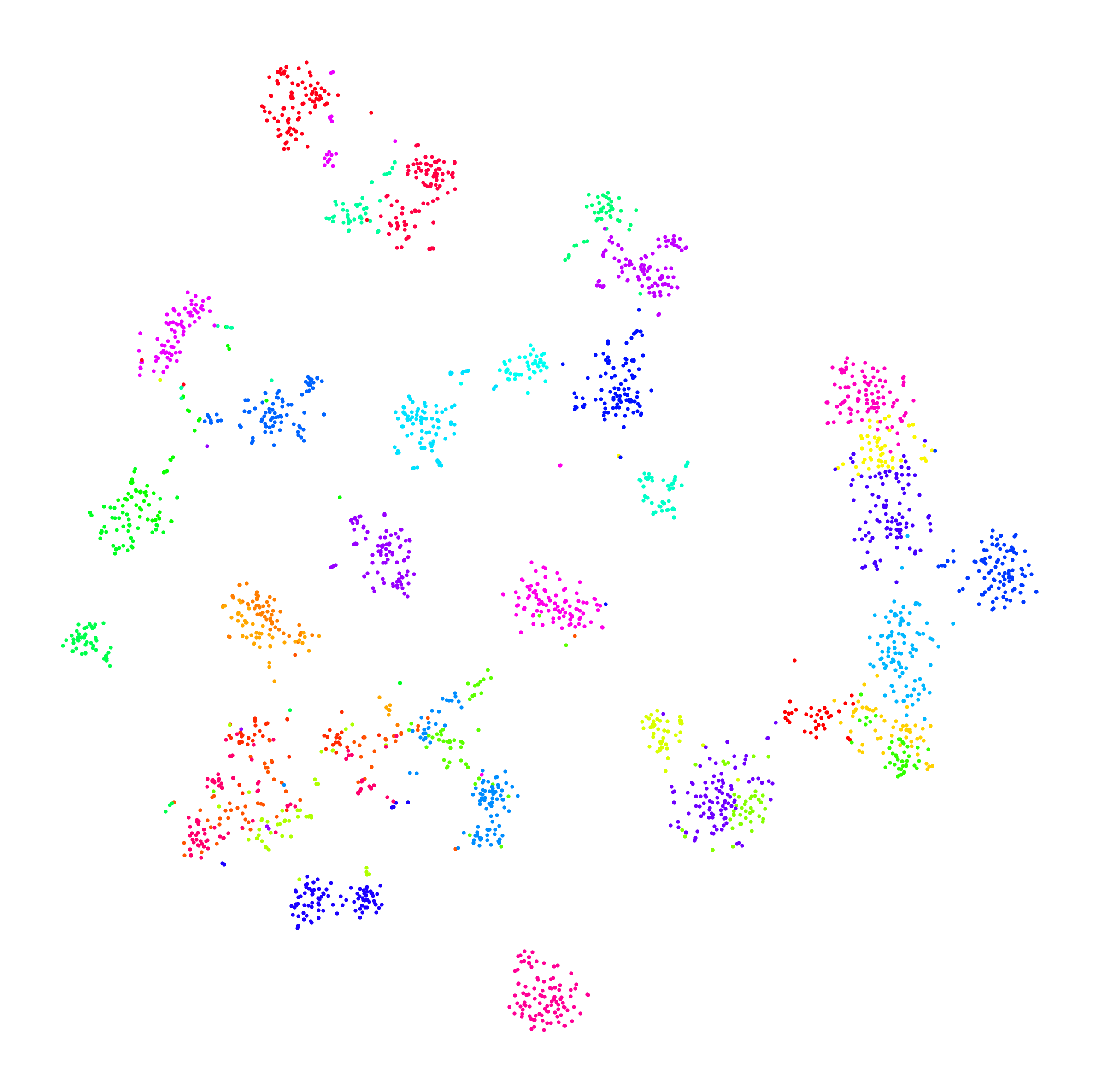}
    \caption{GCD}
    \label{fig:GCD}
  \end{subfigure}
  \begin{subfigure}{.24\linewidth}
    \includegraphics[width=\linewidth]{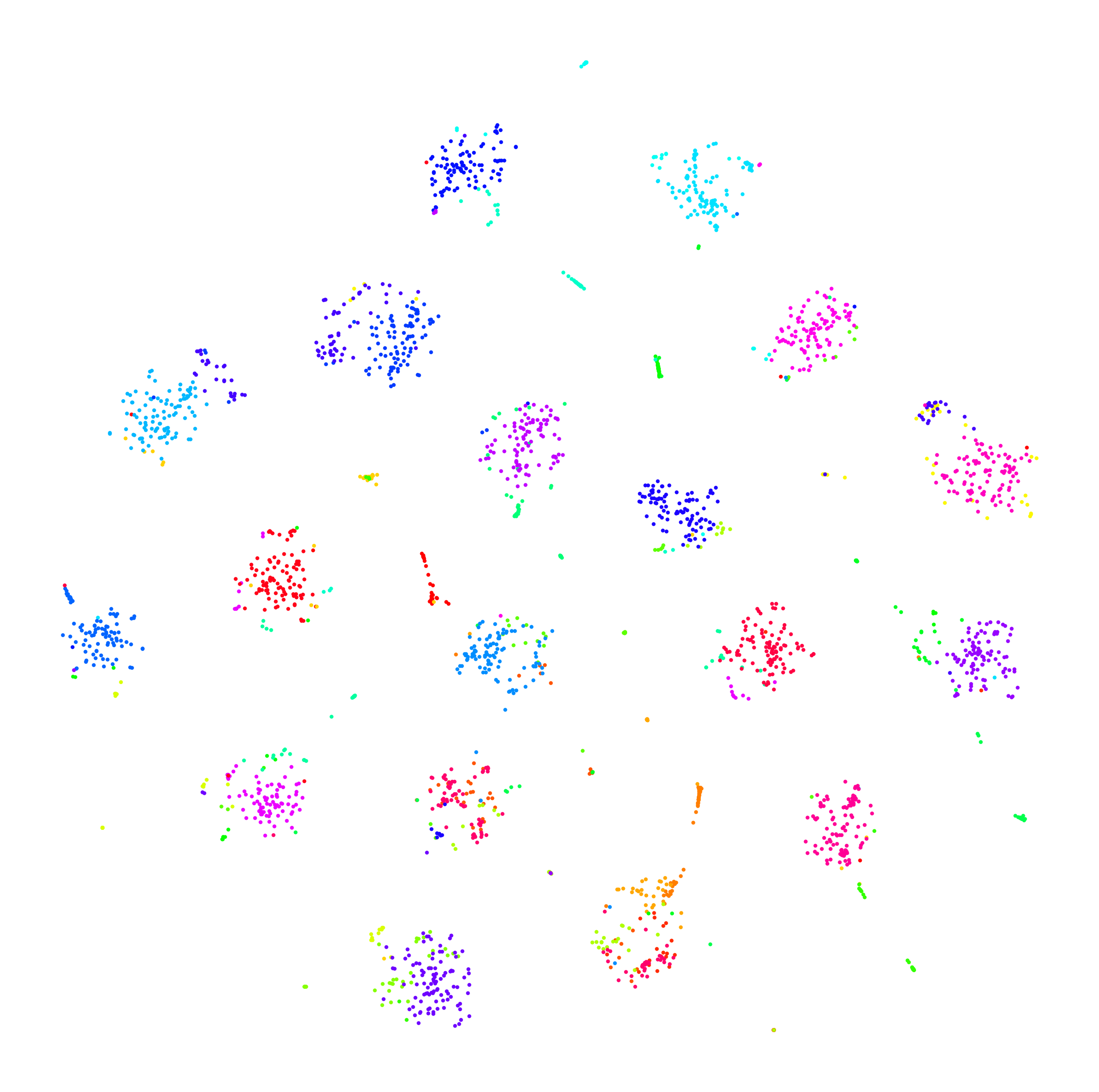}
    \caption{SimGCD}
    \label{fig:SimGCD}
  \end{subfigure}
  \begin{subfigure}{.24\linewidth}
    \includegraphics[width=\linewidth]{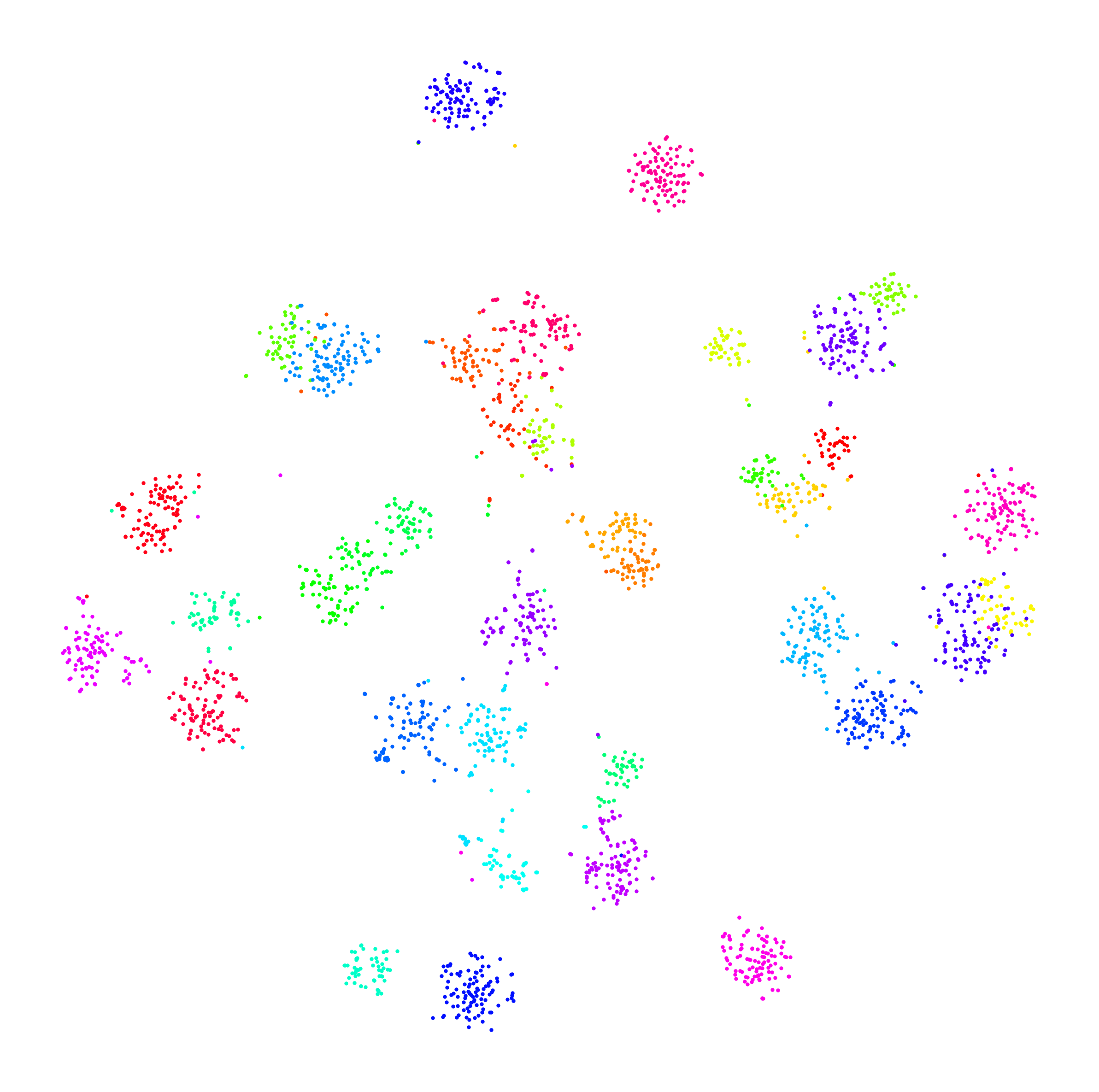}
    \caption{Ours-Image}
    \label{fig:Ours-Image}
  \end{subfigure}
  \begin{subfigure}{.24\linewidth}
    \includegraphics[width=\linewidth]{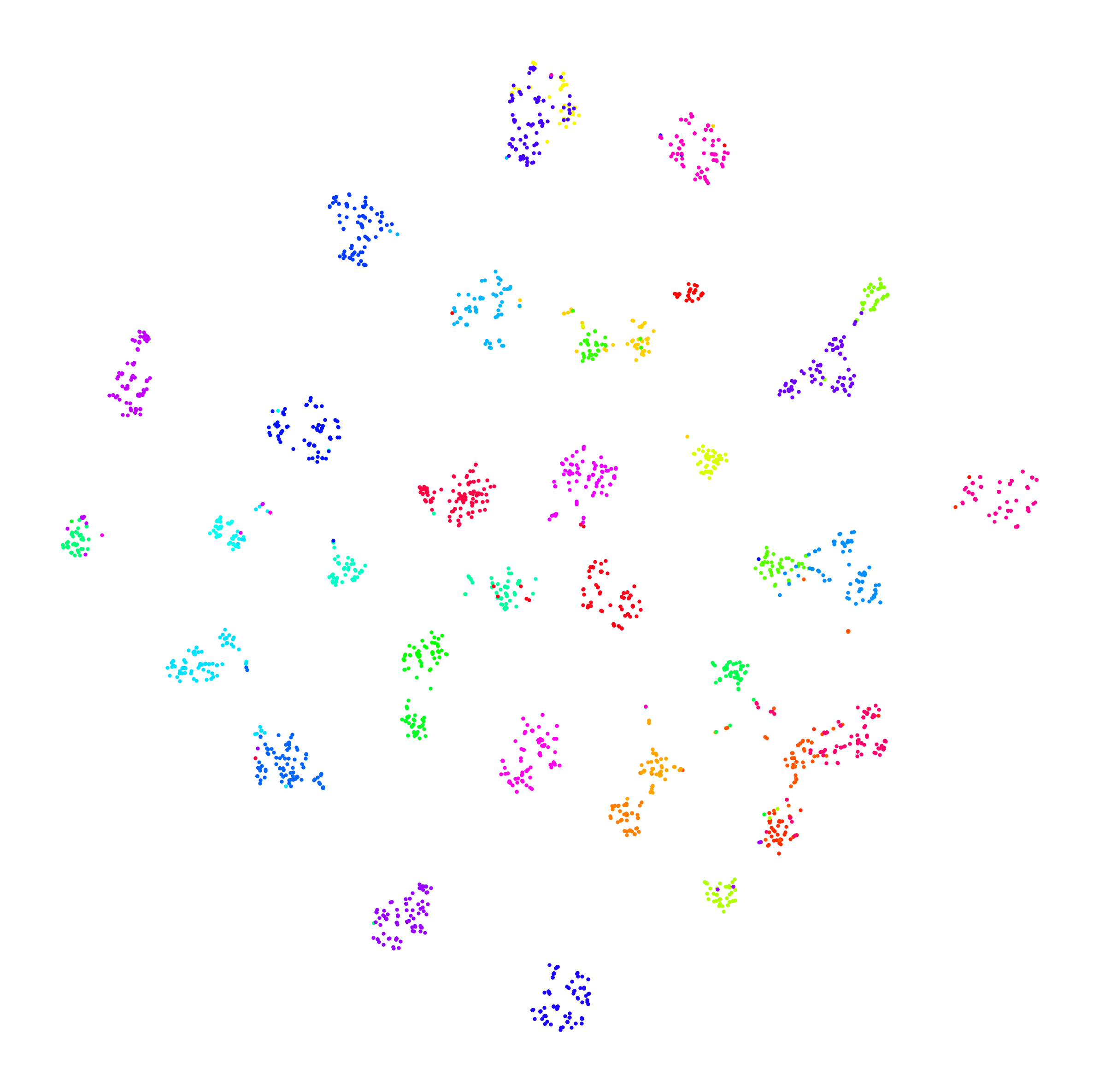}
    \caption{Ours-Text}
    \label{fig:Ours-Text}
  \end{subfigure}
  
  \caption{Visualization of feature distributions of the unlabeled set of the Oxford Pets dataset.}
  \label{fig:feature_distributions}
\end{figure}

To further evaluate the effectiveness of TextGCD, we utilize t-SNE for visualizing feature representations generated by the backbone network. We compare the trained backbones of GCD, SimGCD, and our TextGCD on the Oxford Pets~\cite{pets} dataset. As depicted in Fig.~\ref{fig:feature_distributions} (a), (b), and (c)—all utilizing ViT-B as the backbone—it is evident that TextGCD produces more discriminative feature representations. Moreover, Fig.~\ref{fig:feature_distributions} (d) further validates the effectiveness of TextGCD's textual feature representations.

\section*{C. Limitations}
\noindent \textbf{Dependency on the Visual Lexicon.} 
The effectiveness of our method largely depends on a rich visual lexicon. Although massive category names have been gathered from existing datasets, it might still lack the categories in specific tasks, such as the Herbarium 19 dataset~\cite{herb} designed for botanical research. In such a context, our method may fail to produce informative textual descriptions and thus produce inferior results.

\noindent \textbf{Inheriting Flaws from Foundation Models}. 
Owing to our reliance on LLMs and VLMs for constructing category descriptive texts, our method inevitably inherits biases and drawbacks from these foundational models. For example, as shown in Tab.~\ref{tab:limit_clip}, the CLIP model exhibits suboptimal performance (Zero-Shot) on the FGVC Aircraft dataset~\cite{aircraft}. As a result, our method fails to generate informative text descriptions and cannot obtain improvement compared to the visual-only competitor SimGCD.

\begin{table}[th]
  \caption{Comparison on the FGVC Aircraft dataset. ``Zero Shot'' indicates building the textual classifier with ground-truth names and directly performing classification.}
  \centering
  \scriptsize
  \setlength{\tabcolsep}{10pt}
  \renewcommand{\arraystretch}{0.85}
  \begin{tabular}{lccc}
    \toprule
    Methods & All & Old & New \\
    \midrule
    Zero Shot & 27.0 & 21.3 & 32.3 \\
    \midrule
    GCD & 45.0 & 41.1 & 46.9 \\
    SimGCD & \textbf{54.2} & \textbf{59.1} & 51.8 \\
    Ours & 50.8 & 44.9 & \textbf{53.8} \\
    \bottomrule
  \end{tabular}
  \label{tab:limit_clip}
\end{table}

\section*{D. Discussion and Future Direction}
\subsection*{D.1. Visual Lexicon Selection}
 There is overlap between the datasets used for evaluation and those used to build the Visual Lexicon. Specifically, ImageNet-1K is utilized in both contexts. Our goal is to collect as much object semantics as possible to construct a rich and diverse Visual Lexicon. Given its popularity and the extensive common visual object semantics it contains, which are also prevalent in other datasets, ImageNet-1K's class names are included when constructing the Visual Lexicon.

\subsection*{D.2. Tailored Visual Lexicon}
Since labeled data are available in the GCD task, they can be used to confirm the scope of classified images, especially for fine-grained classification. We create a tailored Visual Lexicon for the Flower102 dataset by employing an LLM (Llama-2-13b) model to filter out tags unrelated to flowers and plants. As shown in Tab.~\ref{tab:evaluation_lexicon}, the tailored Visual Lexicon leads to improved results. In our future work, we plan to construct unique Visual Lexicons for each dataset.

\begin{table}[th]
  \centering
  \scriptsize
  \setlength{\tabcolsep}{10pt}
  \renewcommand{\arraystretch}{1.0}
  \caption{Evaluation about Visual Lexicon on Flower102.}
  \begin{tabular}{lccc}
    \toprule
    Methods & All  & Old  & New \\
    \midrule
    Original Lexicon & 87.2 & 90.7 & 85.4 \\
    Tailored Lexicon & \textbf{88.3} & \textbf{91.4} & \textbf{86.8} \\
    \bottomrule
  \end{tabular}
  \label{tab:evaluation_lexicon}
\end{table}

\subsection*{D.3. Extend to SSL Setting} Our method can be easily extended to a semi-supervised (SSL) setting. Specifically, in a semi-supervised setting, the number of categories and the category names can be considered known. By applying our method to SSL, we can create a Visual Lexicon with ground-truth class names along with attributes obtained by LLMs (such as GPT-3). Based on this tailored Visual Lexicon, our TextGCD can then be applied to the SSL setting.

\subsection*{D.4. Extend to Incremental Setting}
For GCD models to be effectively applied in real-world scenarios, Zhang et al. proposed the Continuous Category Discovery (CCD) task~\cite{GM}, which not only requires the ability to recognize data from incrementally increasing unknown categories but also emphasizes the importance of not forgetting previously known categories. Our method can be applied to the CCD task, as each training stage in CCD is a variant of GCD. The key to this extension is adding a tailored module to prevent forgetting the categories seen in previous training stages, such as the static-dynamic distillation loss in GM~\cite{GM} or the distillation loss in iCaRL~\cite{icarl}. It is essential to maintain the memory of previously learned categories at each training stage. This is the design plan for extending TextGCD to the incremental setting.

\section*{E. Illustration}
\subsection*{E.1. Visual Lexicon}
During the RTG phase, we introduce the category understanding capabilities of LLMs (GPT-3) to expand the Visual Lexicon. According to Tab.7 in the main paper, the introduction of attributes has led to improvements across both generic datasets and fine-grained classification datasets. Some examples from the Visual Lexicon are presented in Tab.~\ref{tab:text_examples}.

\begin{table}[h]
  \centering
  \caption{Examples in the Visual Lexicon.}
  \scriptsize
  \setlength{\tabcolsep}{3pt}
  \renewcommand{\arraystretch}{0.92}
  \begin{tabular}{|l|}
    \hline
    \textbf{Abyssinian cat} \\
    \begin{minipage}[t]{\linewidth}
        \begin{itemize}
            \item which has medium-sized cat
            \item which has short, soft coat
            \item which has distinctive ticked tabby pattern
            \item which has large pointed ears
            \item which is almond-shaped eyes
            \item which has long, lean body
        \end{itemize}
    \end{minipage} \\
    \hline
    \textbf{Bengal cat} \\
    \begin{minipage}[t]{\linewidth}
        \begin{itemize}
            \item which has medium-sized cat with a muscular body
            \item which has thick fur in various shades of brown and black
            \item which has faint horizontal stripes on the neck, face, and legs
            \item which has large, round eyes in shades of gold and green
            \item which has pointed ears
        \end{itemize}
    \end{minipage} \\
    \hline

\textbf{Birman cat} \\
\begin{minipage}[t]{\linewidth}
    \begin{itemize}
        \item which has medium-sized cat
        \item which has long, flowing coat
        \item which has white feet
        \item which has sapphire blue eyes
        \item which has white muzzle
        \item which has medium-length tail
        \item which has long, silky fur
    \end{itemize}
\end{minipage} \\
\hline
\textbf{Bombay cat} \\
\begin{minipage}[t]{\linewidth}
    \begin{itemize}
        \item which has short, dense coat
        \item which has black or black-brown coat with copper highlights
        \item which has large, rounded ears
        \item which is a broad chest and muscular body
        \item which is a short, thick tail
        \item which has four white paws and a white chin
    \end{itemize}
\end{minipage} \\
\hline
\textbf{British Shorthair cat} \\
\begin{minipage}[t]{\linewidth}
    \begin{itemize}
        \item which has short-haired, stocky cat
        \item which has large, round eyes
        \item which has broad head and cheeks
        \item which has short, dense fur
        \item which has short, thick legs
        \item which has short, thick tail
        \item which has thick, dense coat
    \end{itemize}
\end{minipage} \\
\hline
\textbf{Egyptian Mau cat} \\
\begin{minipage}[t]{\linewidth}
    \begin{itemize}
        \item which has short-haired cat
        \item which has spotted or marbled coat
        \item which has large eyes
        \item which has slender body
        \item which has muscular hindquarters
        \item which has long, thin tail
    \end{itemize}
\end{minipage} \\
\hline
  \end{tabular}
  \label{tab:text_examples}
\end{table}

\subsection*{E.2. Categorical Descriptive Text}
\label{sec:tags_attributes}
\begin{figure}[!htbp]
  \centering
  \begin{subfigure}{0.87\linewidth}
    \includegraphics[width=\linewidth]{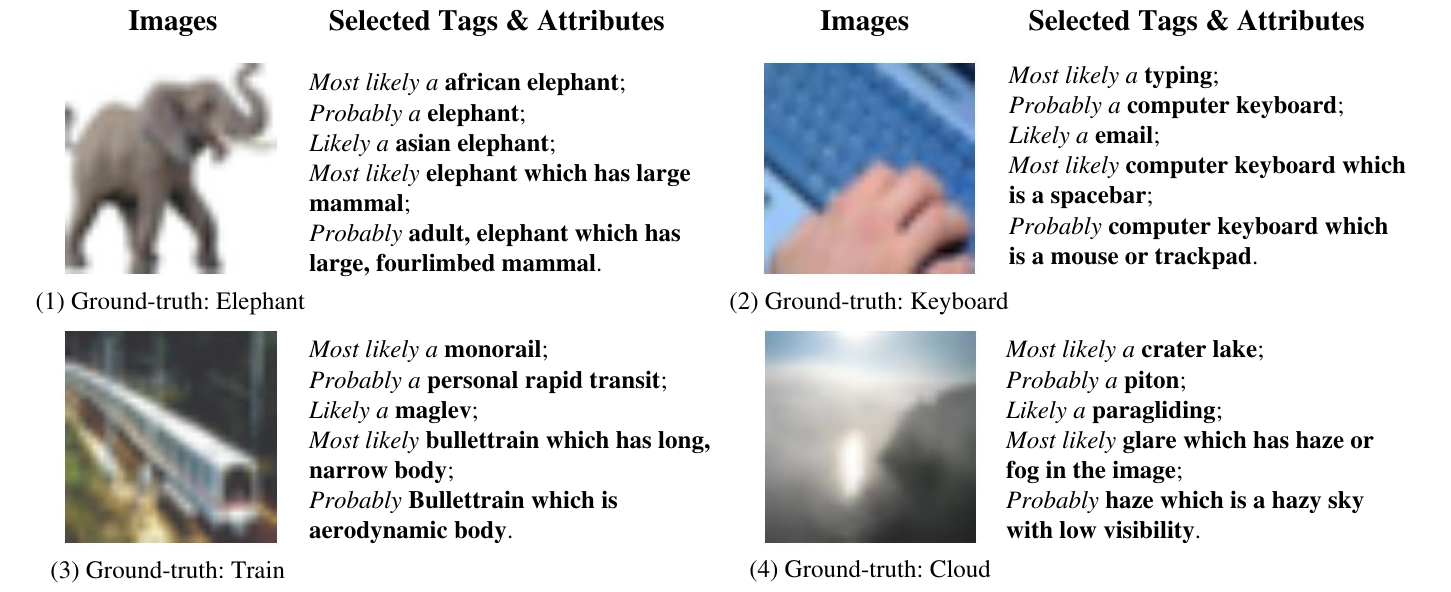}
    \caption{Examples in CIFAR-100}
  \end{subfigure}
  \begin{subfigure}{0.87\linewidth}
    \includegraphics[width=\linewidth]{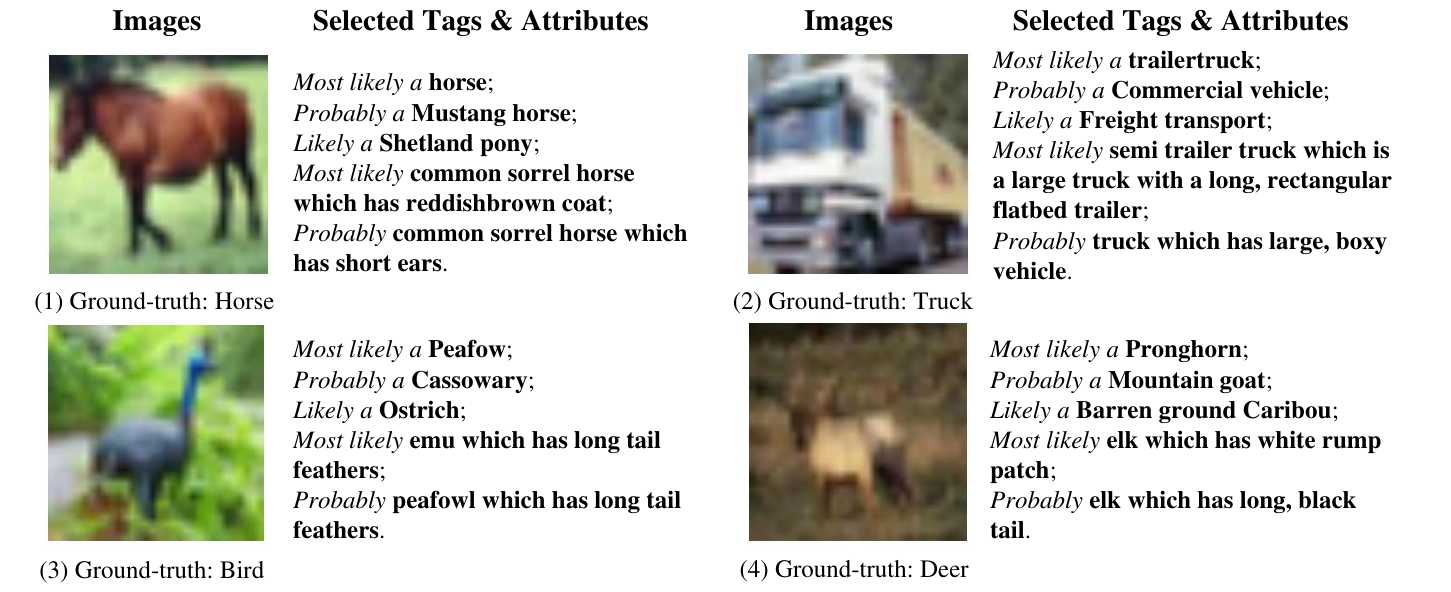}
    \caption{Examples in CIFAR-10}
  \end{subfigure}
  \begin{subfigure}{0.87\linewidth}
    \includegraphics[width=\linewidth]{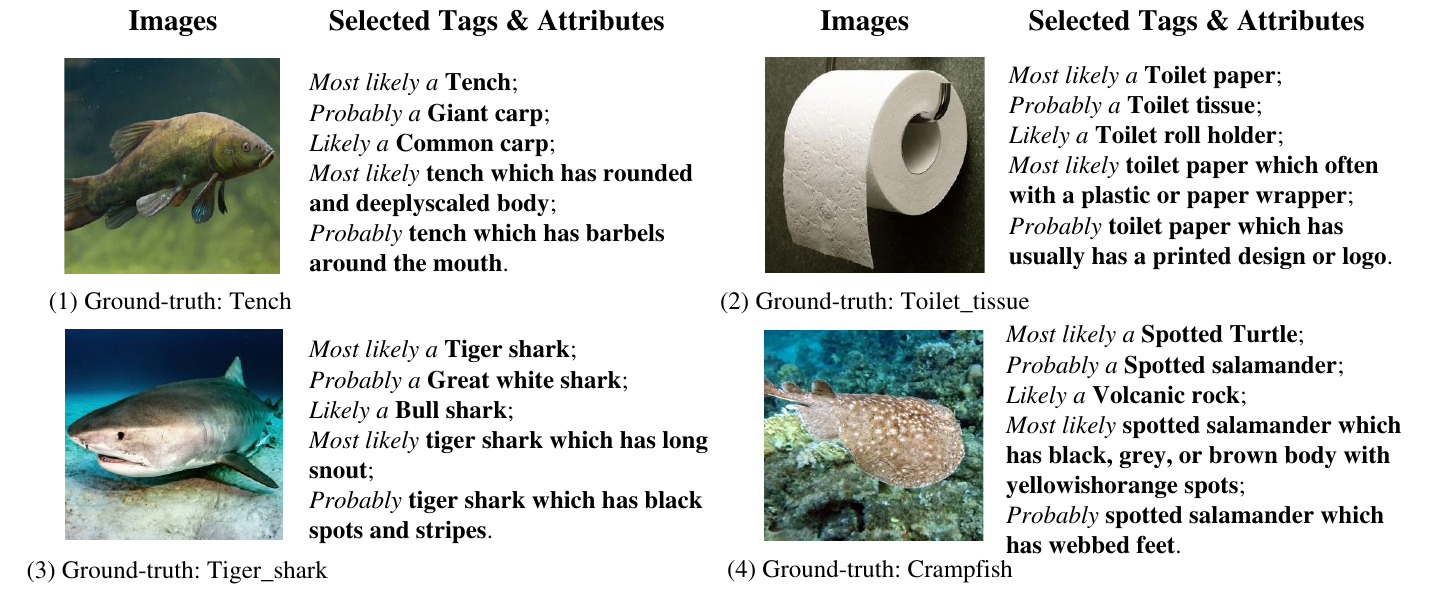}
    \caption{Examples in ImageNet}
  \end{subfigure}
  
  \caption{Generated categorical descriptive text in the RTG phase: representative examples in generic datasets.}
  \label{fig:case_generic}
\end{figure}

\begin{figure}[!htbp]
  \centering
  \begin{subfigure}{0.87\linewidth}
    \includegraphics[width=\linewidth]{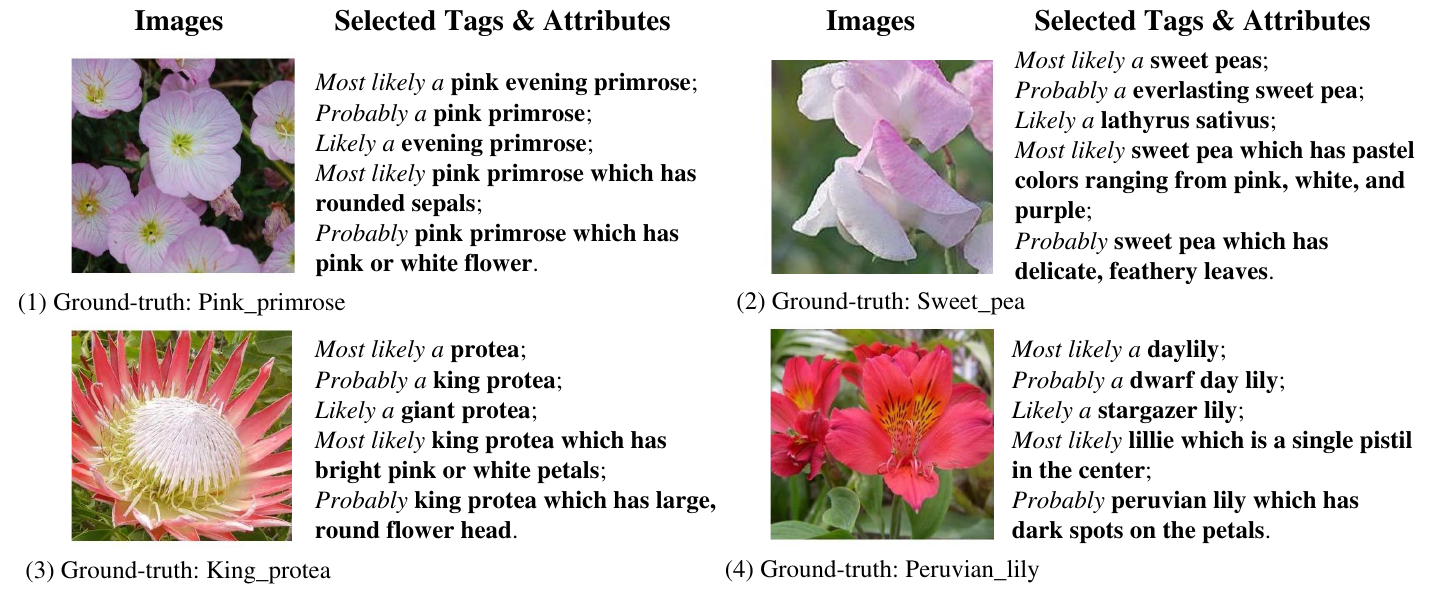}
    \caption{Examples in Flowers102}
  \end{subfigure}
  \begin{subfigure}{0.87\linewidth}
    \includegraphics[width=\linewidth]{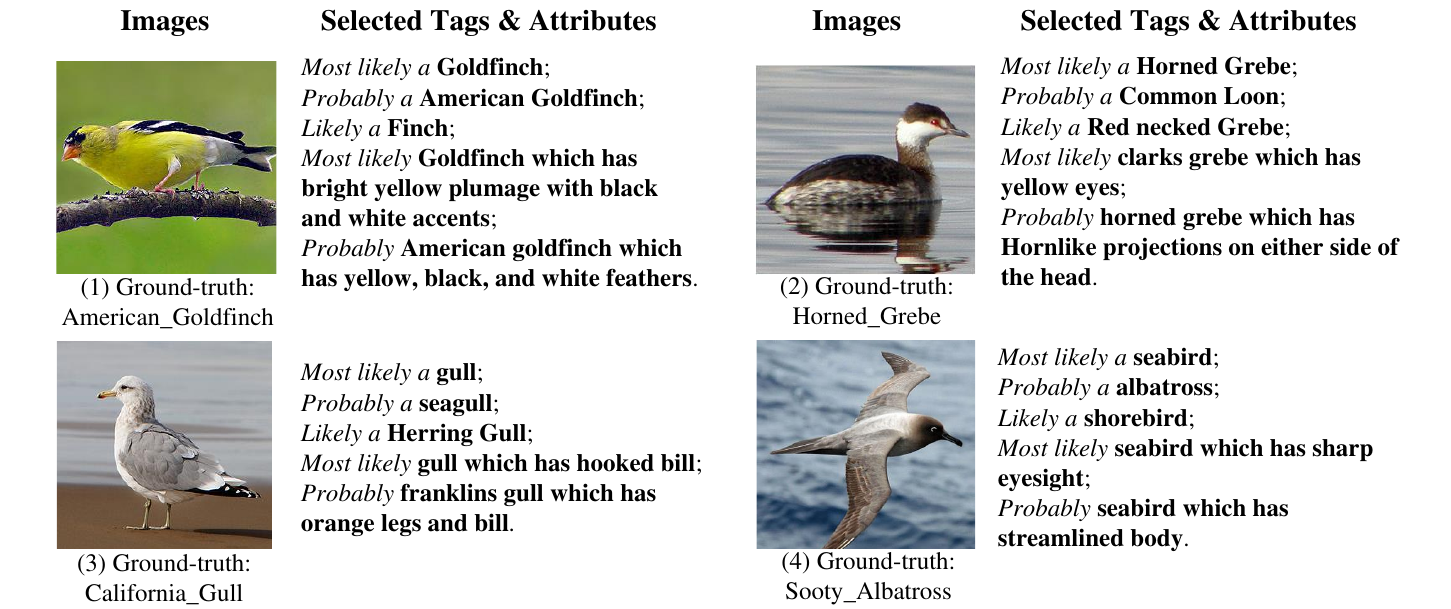}
    \caption{Examples in CUB}
  \end{subfigure}
  \begin{subfigure}{0.87\linewidth}
    \includegraphics[width=\linewidth]{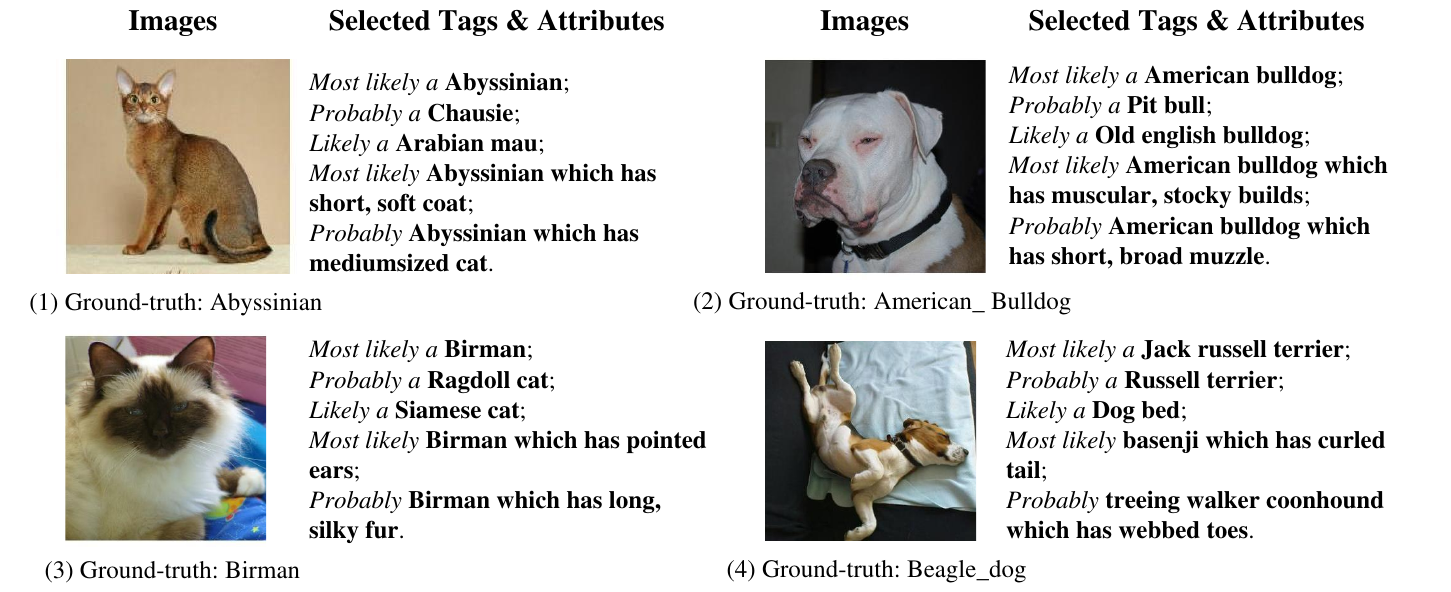}
    \caption{Examples in Oxford Pets}
  \end{subfigure}
  \begin{subfigure}{0.87\linewidth}
    \includegraphics[width=\linewidth]{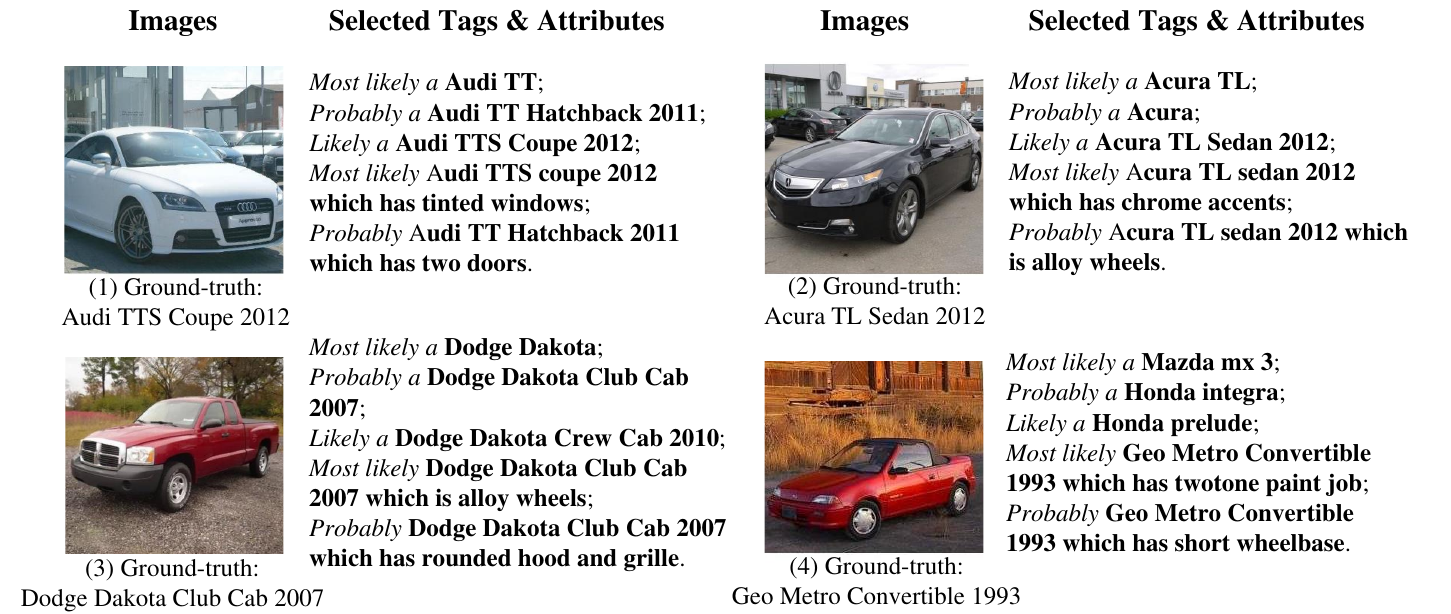}
    \caption{Examples in Stanford Cars}
  \end{subfigure}

  \caption{Generated categorical descriptive text in the RTG phase: representative examples in fine-grained datasets.}
  \label{fig:case_finegrained}
\end{figure}

In Fig.~\ref{fig:case_generic} and Fig.~\ref{fig:case_finegrained}, we showcase examples of descriptive text generated during the RTG phase. For images with distinct features, the generated text accurately describes the categories, as illustrated in Fig.~\ref{fig:case_generic} (a.1 2), Fig.~\ref{fig:case_generic} (c.1,2,3), and Fig.~\ref{fig:case_finegrained} (a.1,2; b.1,2). In some cases, the descriptive text displays categories closely related to the ground truth, such as ``monorail'' and ``maglev'' for ``Train'' in Fig.~\ref{fig:case_generic} (a.3), and ``trailer truck'' and ``commercial vehicle'' for 'Truck' in Fig.~\ref{fig:case_generic} (b.2). Additionally, the text sometimes presents subcategories of the ground truth, like ``Peafowl'' and ``Cassowary'' for ``Bird'' in Fig.~\ref{fig:case_generic} (b.3). This textual information aids in achieving precise category differentiation, facilitating mutual enhancement with visual modalities during the co-teaching phase.

However, there are instances where the text generates incorrect category descriptions, such as in Fig.~\ref{fig:case_generic} (a.4), (b.4), (c.4) and Fig.~\ref{fig:case_finegrained} (a.4), (c.4), (d.4). Yet, these category descriptive texts reflect certain features of the true categories. For example, ``fog'' and ``sky'' in Fig.~\ref{fig:case_generic} (a.4) are relevant to ``Cloud''; ``Pronghorn'', ``Mountain goat'', and ``deer'' share similar appearances in Fig.~\ref{fig:case_generic} (b.4); ``Peruvian lily'', ``daylily'', and ``dwarf day lily'' in Fig.~\ref{fig:case_finegrained} (a.4) belong to the lily family. On one hand, this association promotes mutual learning during the co-teaching process. On the other hand, we utilize a soft voting mechanism to combine the category determinations from textual and visual modalities, jointly deciding on category divisions. This approach helps to reduce misjudgments made by relying on a single modality.

\clearpage  

%
%

%% file: main.bbl
\begin{thebibliography}{10}
\providecommand{\url}[1]{\texttt{#1}}
\providecommand{\urlprefix}{URL }
\providecommand{\doi}[1]{https://doi.org/#1}

\bibitem{assran2022masked}
Assran, M., Caron, M., Misra, I., Bojanowski, P., Bordes, F., Vincent, P., Joulin, A., Rabbat, M., Ballas, N.: Masked siamese networks for label-efficient learning. In: European Conference on Computer Vision (2022)

\bibitem{lens}
Berrios, W., Mittal, G., Thrush, T., Kiela, D., Singh, A.: Towards language models that can see: Computer vision through the lens of natural language. arXiv preprint arXiv:2306.16410  (2023)

\bibitem{co-training}
Blum, A., Mitchell, T.: Combining labeled and unlabeled data with co-training. In: Proceedings of the eleventh annual conference on Computational learning theory. pp. 92--100 (1998)

\bibitem{food}
Bossard, L., Guillaumin, M., Van~Gool, L.: Food-101--mining discriminative components with random forests. In: European Conference on Computer Vision (2014)

\bibitem{gpt3}
Brown, T., Mann, B., Ryder, N., Subbiah, M., Kaplan, J.D., Dhariwal, P., Neelakantan, A., Shyam, P., Sastry, G., Askell, A., et~al.: Language models are few-shot learners. Advances in Neural Information Processing Systems  (2020)

\bibitem{dino}
Caron, M., Touvron, H., Misra, I., J{\'e}gou, H., Mairal, J., Bojanowski, P., Joulin, A.: Emerging properties in self-supervised vision transformers. In: Proceedings of the IEEE/CVF International Conference on Computer Vision (2021)

\bibitem{wild}
Cimpoi, M., Maji, S., Kokkinos, I., Mohamed, S., Vedaldi, A.: Describing textures in the wild. In: Proceedings of the IEEE/CVF Conference on Computer Vision and Pattern Recognition (2014)

\bibitem{imagenet}
Deng, J., Dong, W., Socher, R., Li, L.J., Li, K., Fei-Fei, L.: Imagenet: A large-scale hierarchical image database. In: Proceedings of the IEEE/CVF International Conference on Computer Vision (2009)

\bibitem{vit}
Dosovitskiy, A., Beyer, L., Kolesnikov, A., Weissenborn, D., Zhai, X., Unterthiner, T., Dehghani, M., Minderer, M., Heigold, G., Gelly, S., et~al.: An image is worth 16x16 words: Transformers for image recognition at scale. arXiv preprint arXiv:2010.11929  (2020)

\bibitem{fini2021unified}
Fini, E., Sangineto, E., Lathuili{\`e}re, S., Zhong, Z., Nabi, M., Ricci, E.: A unified objective for novel class discovery. In: Proceedings of the IEEE/CVF International Conference on Computer Vision (2021)

\bibitem{lvis}
Gupta, A., Dollar, P., Girshick, R.: Lvis: A dataset for large vocabulary instance segmentation. In: Proceedings of the IEEE/CVF Conference on Computer Vision and Pattern Recognition (2019)

\bibitem{coteaching}
Han, B., Yao, Q., Yu, X., Niu, G., Xu, M., Hu, W., Tsang, I., Sugiyama, M.: Co-teaching: Robust training of deep neural networks with extremely noisy labels. Advances in Neural Information Processing Systems  (2018)

\bibitem{han2021autonovel}
Han, K., Rebuffi, S.A., Ehrhardt, S., Vedaldi, A., Zisserman, A.: Autonovel: Automatically discovering and learning novel visual categories. IEEE Transactions on Pattern Analysis and Machine Intelligence  (2021)

\bibitem{han2019learning}
Han, K., Vedaldi, A., Zisserman, A.: Learning to discover novel visual categories via deep transfer clustering. In: Proceedings of the IEEE/CVF International Conference on Computer Vision (2019)

\bibitem{scars}
Krause, J., Stark, M., Deng, J., Fei-Fei, L.: 3d object representations for fine-grained categorization. In: Proceedings of the IEEE/CVF International Conference on Computer Vision Workshops (2013)

\bibitem{3dobject}
Krause, J., Stark, M., Deng, J., Fei-Fei, L.: 3d object representations for fine-grained categorization. In: Proceedings of the IEEE/CVF International Conference on Computer Vision Workshops (2013)

\bibitem{visualgenome}
Krishna, R., Zhu, Y., Groth, O., Johnson, J., Hata, K., Kravitz, J., Chen, S., Kalantidis, Y., Li, L.J., Shamma, D.A., et~al.: Visual genome: Connecting language and vision using crowdsourced dense image annotations. International Journal of Computer Vision  (2017)

\bibitem{cifar}
Krizhevsky, A., Hinton, G., et~al.: Learning multiple layers of features from tiny images. Technical Report  (2009)

\bibitem{openimages}
Kuznetsova, A., Rom, H., Alldrin, N., Uijlings, J., Krasin, I., Pont-Tuset, J., Kamali, S., Popov, S., Malloci, M., Kolesnikov, A., et~al.: The open images dataset v4: Unified image classification, object detection, and visual relationship detection at scale. International Journal of Computer Vision  (2020)

\bibitem{Caltech101}
Li, F.F., Andreeto, M., Ranzato, M., Perona, P.: Caltech 101 (2022)

\bibitem{blip2}
Li, J., Li, D., Savarese, S., Hoi, S.: Blip-2: Bootstrapping language-image pre-training with frozen image encoders and large language models. arXiv preprint arXiv:2301.12597  (2023)

\bibitem{blip}
Li, J., Li, D., Xiong, C., Hoi, S.: Blip: Bootstrapping language-image pre-training for unified vision-language understanding and generation. In: International Conference on Machine Learning (2022)

\bibitem{flip}
Li, Y., Fan, H., Hu, R., Feichtenhofer, C., He, K.: Scaling language-image pre-training via masking. In: Proceedings of the IEEE/CVF Conference on Computer Vision and Pattern Recognition (2023)

\bibitem{coco}
Lin, T.Y., Maire, M., Belongie, S., Hays, J., Perona, P., Ramanan, D., Doll{\'a}r, P., Zitnick, C.L.: Microsoft coco: Common objects in context. In: European Conference on Computer Vision (2014)

\bibitem{kmeans}
MacQueen, J., et~al.: Some methods for classification and analysis of multivariate observations. In: Proceedings of the Fifth Berkeley Symposium on Mathematical Statistics and Probability (1967)

\bibitem{aircraft}
Maji, S., Rahtu, E., Kannala, J., Blaschko, M., Vedaldi, A.: Fine-grained visual classification of aircraft. arXiv preprint arXiv:1306.5151  (2013)

\bibitem{vlm_for_cls1}
Menon, S., Vondrick, C.: Visual classification via description from large language models. arXiv preprint arXiv:2210.07183  (2022)

\bibitem{flowers}
Nilsback, M.E., Zisserman, A.: Automated flower classification over a large number of classes. In: 2008 Sixth Indian Conference on Computer Vision, Graphics \& Image Processing (2008)

\bibitem{vlm_for_cls3}
Novack, Z., McAuley, J., Lipton, Z.C., Garg, S.: Chils: Zero-shot image classification with hierarchical label sets. In: International Conference on Machine Learning. PMLR (2023)

\bibitem{clipgcd}
Ouldnoughi, R., Kuo, C.W., Kira, Z.: Clip-gcd: Simple language guided generalized category discovery. arXiv preprint arXiv:2305.10420  (2023)

\bibitem{pets}
Parkhi, O.M., Vedaldi, A., Zisserman, A., Jawahar, C.: Cats and dogs. In: Proceedings of the IEEE/CVF Conference on Computer Vision and Pattern Recognition (2012)

\bibitem{dccl}
Pu, N., Zhong, Z., Sebe, N.: Dynamic conceptional contrastive learning for generalized category discovery. In: Proceedings of the IEEE/CVF Conference on Computer Vision and Pattern Recognition (2023)

\bibitem{clip}
Radford, A., Kim, J.W., Hallacy, C., Ramesh, A., Goh, G., Agarwal, S., Sastry, G., Askell, A., Mishkin, P., Clark, J., et~al.: Learning transferable visual models from natural language supervision. In: International Conference on Machine Learning (2021)

\bibitem{icarl}
Rebuffi, S.A., Kolesnikov, A., Sperl, G., Lampert, C.H.: icarl: Incremental classifier and representation learning. In: Proceedings of the IEEE conference on Computer Vision and Pattern Recognition. pp. 2001--2010 (2017)

\bibitem{coteaching_for_MDA}
Roy, S., Krivosheev, E., Zhong, Z., Sebe, N., Ricci, E.: Curriculum graph co-teaching for multi-target domain adaptation. In: Proceedings of the IEEE/CVF conference on computer vision and pattern recognition (2021)

\bibitem{roy2022class}
Roy, S., Liu, M., Zhong, Z., Sebe, N., Ricci, E.: Class-incremental novel class discovery. In: European Conference on Computer Vision (2022)

\bibitem{imagenetlarge}
Russakovsky, O., Deng, J., Su, H., Krause, J., Satheesh, S., Ma, S., Huang, Z., Karpathy, A., Khosla, A., Bernstein, M., et~al.: Imagenet large scale visual recognition challenge. International Journal of Computer Vision  (2015)

\bibitem{sloutsky2010perceptual}
Sloutsky, V.M.: From perceptual categories to concepts: What develops? Cognitive science  (2010)

\bibitem{herb}
Tan, K.C., Liu, Y., Ambrose, B., Tulig, M., Belongie, S.: The herbarium challenge 2019 dataset. arXiv preprint arXiv:1906.05372  (2019)

\bibitem{gcd}
Vaze, S., Han, K., Vedaldi, A., Zisserman, A.: Generalized category discovery. In: Proceedings of the IEEE/CVF Conference on Computer Vision and Pattern Recognition (2022)

\bibitem{cub}
Wah, C., Branson, S., Welinder, P., Perona, P., Belongie, S.: The caltech-ucsd birds-200-2011 dataset. Computation \& Neural Systems Technical Report  (2011)

\bibitem{sptnet}
Wang, H., Vaze, S., Han, K.: Sptnet: An efficient alternative framework for generalized category discovery with spatial prompt tuning. In: The Twelfth International Conference on Learning Representations (2023)

\bibitem{simgcd}
Wen, X., Zhao, B., Qi, X.: Parametric classification for generalized category discovery: A baseline study. In: Proceedings of the IEEE/CVF International Conference on Computer Vision (2023)

\bibitem{sundatabase}
Xiao, J., Hays, J., Ehinger, K.A., Oliva, A., Torralba, A.: Sun database: Large-scale scene recognition from abbey to zoo. In: Proceedings of the IEEE/CVF Conference on Computer Vision and Pattern Recognition (2010)

\bibitem{coteaching_for_reid1}
Yang, F., Li, K., Zhong, Z., Luo, Z., Sun, X., Cheng, H., Guo, X., Huang, F., Ji, R., Li, S.: Asymmetric co-teaching for unsupervised cross-domain person re-identification. In: Proceedings of the AAAI conference on artificial intelligence (2020)

\bibitem{vlm_for_cls2}
Yang, Y., Panagopoulou, A., Zhou, S., Jin, D., Callison-Burch, C., Yatskar, M.: Language in a bottle: Language model guided concept bottlenecks for interpretable image classification. In: Proceedings of the IEEE/CVF Conference on Computer Vision and Pattern Recognition (2023)

\bibitem{coca}
Yu, J., Wang, Z., Vasudevan, V., Yeung, L., Seyedhosseini, M., Wu, Y.: Coca: Contrastive captioners are image-text foundation models. arXiv preprint arXiv:2205.01917  (2022)

\bibitem{coteaching_for_OMO}
Yuan, Y., Chen, C.S., Liu, Z., Neiswanger, W., Liu, X.S.: Importance-aware co-teaching for offline model-based optimization. Advances in Neural Information Processing Systems  (2024)

\bibitem{promptcal}
Zhang, S., Khan, S., Shen, Z., Naseer, M., Chen, G., Khan, F.S.: Promptcal: Contrastive affinity learning via auxiliary prompts for generalized novel category discovery. In: Proceedings of the IEEE/CVF Conference on Computer Vision and Pattern Recognition (2023)

\bibitem{GM}
Zhang, X., Jiang, J., Feng, Y., Wu, Z.F., Zhao, X., Wan, H., Tang, M., Jin, R., Gao, Y.: Grow and merge: A unified framework for continuous categories discovery. Advances in Neural Information Processing Systems  \textbf{35},  27455--27468 (2022)

\bibitem{gpc}
Zhao, B., Wen, X., Han, K.: Learning semi-supervised gaussian mixture models for generalized category discovery. arXiv preprint arXiv:2305.06144  (2023)

\bibitem{zhong2021neighborhood}
Zhong, Z., Fini, E., Roy, S., Luo, Z., Ricci, E., Sebe, N.: Neighborhood contrastive learning for novel class discovery. In: Proceedings of the IEEE/CVF conference on computer vision and pattern recognition (2021)

\bibitem{zhong2021openmix}
Zhong, Z., Zhu, L., Luo, Z., Li, S., Yang, Y., Sebe, N.: Openmix: Reviving known knowledge for discovering novel visual categories in an open world. In: Proceedings of the IEEE/CVF Conference on Computer Vision and Pattern Recognition (2021)

\end{thebibliography}
